\documentclass[lettersize,journal]{IEEEtran}
\usepackage{amsmath,amsfonts}
\usepackage{algorithmic}
\usepackage{algorithm}
\usepackage{array}
\usepackage[caption=false,font=normalsize,labelfont=sf,textfont=sf]{subfig}
\usepackage{textcomp}
\usepackage{stfloats}
\usepackage{url}
\usepackage{verbatim}
\usepackage{graphicx}
\usepackage{cite}
\usepackage{cite}
\usepackage{booktabs}
\usepackage{multirow}
\usepackage{graphicx} 
\usepackage{hyperref}
\usepackage{makecell}

\usepackage{color}
\usepackage[table]{xcolor} 

\begin{document}

\title{Semi-supervised Semantic Segmentation for Remote Sensing Images via Multi-scale Uncertainty Consistency and Cross-Teacher-Student Attention}
\author{\IEEEauthorblockN{Shanwen Wang, Xin Sun,~\IEEEmembership{Senior member,~IEEE}, Changrui Chen, Danfeng Hong,~\IEEEmembership{Senior member,~IEEE}, Jungong Han,~\IEEEmembership{Senior member,~IEEE}
}

\thanks{S. Wang and X. Sun are with Faculty of Data Science, City University of Macau, 999078, SAR Macao, China (Corresponding author: sunxin1984@ieee.org). C. Chen is with WMG, University of Warwick, UK. (Corresponding author: geoffreychen777@gmail.com) D. Hong is with the Aerospace Information Research Institute, Chinese Academy of Sciences, Beijing 100094, China, and also with the School of Electronic, Electrical and Communication Engineering, University of Chinese Academy of Sciences, Beijing 100049, China. J. Han is with Department of Automation, Tsinghua University, Beijing, China. 
}
\thanks{This work is supported by the Science and Technology Development Fund, Macao SAR No.0006/2024/RIA1 and National Natural Science Foundation of China under Project No. 61971388 and 62376237.}
}

\markboth{Journal of \LaTeX\ Class Files,~Vol.~14, No.~8, August~2024}%
{Shell \MakeLowercase{\textit{et al.}}: A Sample Article Using IEEEtran.cls for IEEE Journals}


\maketitle

\begin{abstract}
Semi-supervised learning offers an appealing solution for remote sensing (RS) image segmentation to relieve the burden of labor-intensive pixel-level labeling. However, RS images pose unique challenges, including rich multi-scale features and high inter-class similarity. To address these problems, this paper proposes a novel semi-supervised Multi-Scale Uncertainty and Cross-Teacher-Student Attention (MUCA) model for RS image semantic segmentation tasks. Specifically, MUCA constrains the consistency among feature maps at different layers of the network by introducing a multi-scale uncertainty consistency regularization. It improves the multi-scale learning capability of semi-supervised algorithms on unlabeled data. Additionally, MUCA utilizes a Cross-Teacher-Student attention mechanism to guide the student network, guiding the student network to construct more discriminative feature representations through complementary features from the teacher network. This design effectively integrates weak and strong augmentations (WA and SA) to further boost segmentation performance. To verify the effectiveness of our model, we conduct extensive experiments on ISPRS-Potsdam and LoveDA datasets. The experimental results show the superiority of our method over state-of-the-art semi-supervised methods. Notably, our model excels in distinguishing highly similar objects, showcasing its potential for advancing semi-supervised RS image segmentation tasks. 
\end{abstract}


\section{Introduction}
\IEEEPARstart{D}{eep} learning techniques on sensing images semantic segmentation \cite{he2022swin,hong2024spectralgpt} provide promising solutions for disaster prevention \cite{sun2020deep}, land-use surveillance\cite{xu2021modular}, environment protection \cite{10750822}, and urban planning \cite{zhu2017deep}. The increasing number of Earth-observation satellites makes a large amount of raw remote sensing (RS) images be continuously captured, which amplifies the advantage of deep learning methods\cite{wen2021change}. However, it is a time-consuming and laborious task to manually label a large amount of RS data into a large number of categories with the complexity of labeling rules \cite{xin2024confidence}. Therefore, semi-supervised learning draws the attention of the RS communities, because it only uses a small number of labeled samples and a large amount of unlabeled \cite{huang2024decouple}. 

Several semi-supervised semantic segmentation methods have been conducted for natural images, which are mainly divided into teacher-student consistency\cite{tarvainen2017mean}, feature perturbation consistency\cite{sohn2020fixmatch}, and self-training pseudo-labeling\cite{Wang_2024_CVPR}. Self-training pseudo-labeling methods often experience performance degradation due to incorrect pseudo-labels generated in the early stages, which can lead to error propagation. Feature perturbation consistency approaches rely heavily on selecting appropriate perturbations and balanced labeled data, making them particularly sensitive to data imbalance. Among these, the teacher-student consistency framework stands out as the most stable and is especially well-suited for tasks that demand high model complexity \cite{pelaez2023survey}. Recent works \cite{zhang2023pseudo, huang2024decouple} have introduced semi-supervised segmentation into the field of RS image analysis, meanwhile, encountered some domain-specific problems \cite{jin2024dynamic}, e.g., rich multi-scale information and high inter-class similarities. Such inherent domain gaps between natural and RS images pose great challenges for semi-supervised RS image segmentation. 

 \begin{figure}[!t]
\centering
\includegraphics[width=3.5in]{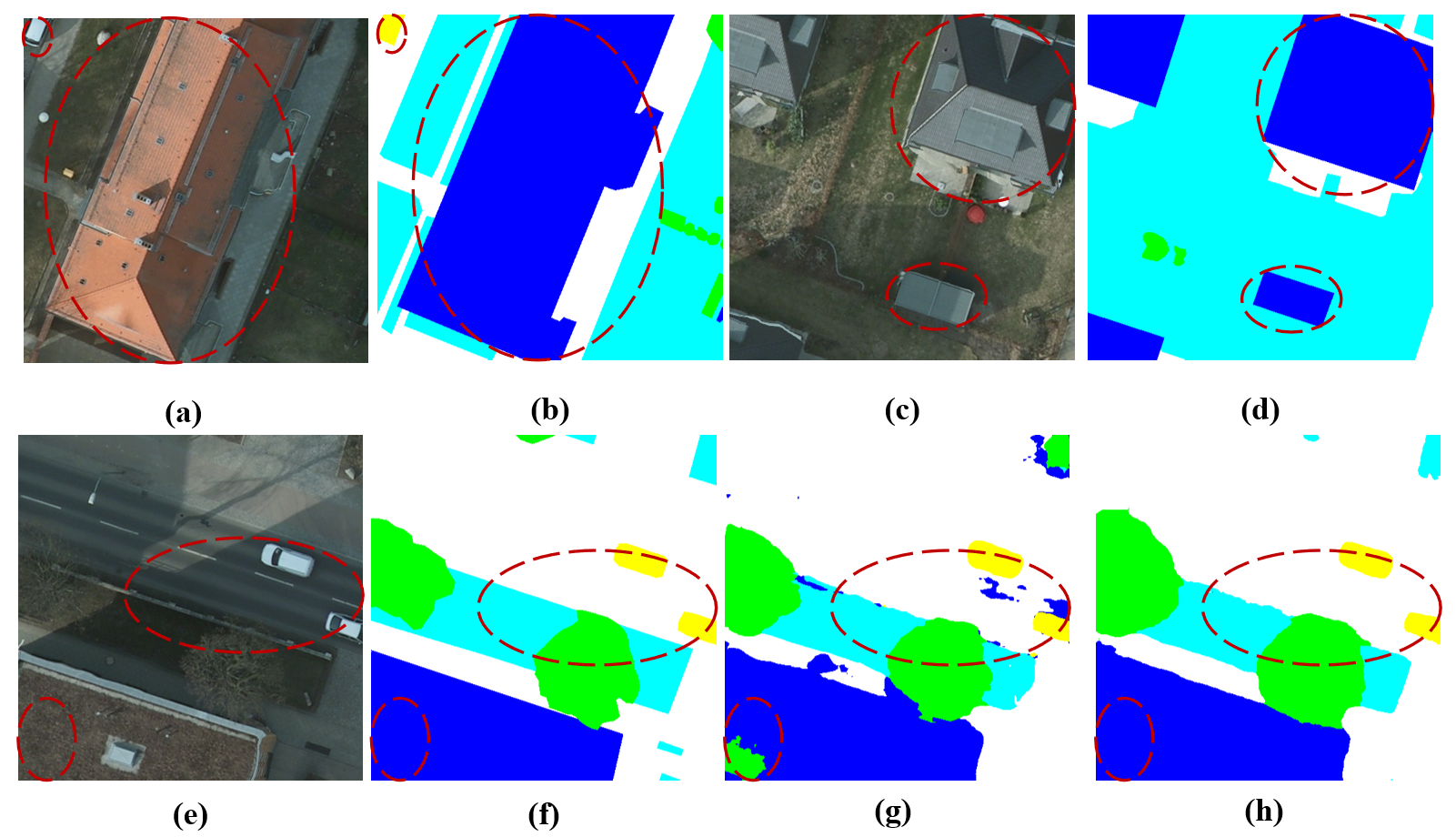}
\vspace{-0.5cm}
\caption{(a) and (b) Image from ISPRS-Potsdam dataset and its Ground Truth. (c) and (d) Image from ISPRS-Potsdam dataset and its  Ground Truth. (e - h) Image from ISPRS-Potsdam dataset, Ground Truth, and results from Unimatch and Ours.}
\label{fg1}
\vspace{-0.5cm}
\end{figure}

\begin{enumerate}
  \item Rich Multi-scale Information\cite{cai2024poly}: Objects on the RS image exhibit a wide range of scales, from expansive objects (e.g., Building) to small entities (e.g., Car). As shown in Fig. \ref{fg1}(a) and Fig. \ref{fg1}(b), the yellow regions represent cars while the blue regions denote buildings, demonstrating the significant size variations across different categories in RS images. Furthermore, objects within the same category exhibit substantial size disparities, as exemplified by the buildings illustrated in Fig. \ref{fg1}(c) and Fig. \ref{fg1}(d).
  

  \item High Inter-class Similarities\cite{zhang2022artificial}: Objects from different categories appear to have high visual similarity and are usually intertwined on the RS images. For example, as shown in. Fig. \ref{fg1}(g), the boundary regions of Low vegetation, Tree, Car, and Building are misclassified with Unimatch \cite{yang2023revisiting}.
\end{enumerate}

 
 To address the above challenges, we propose a new semi-supervised semantic segmentation model for RS images based on Multi-Scale Uncertainty and Cross-Teacher-Student Attention (MUCA). (1) Firstly, we propose for the first time a Multi-Scale Uncertainty Consistency module (MSUC). The module constrains the consistency among feature maps at different layers of the network model, which improves the multi-scale learning capability of semi-supervised algorithms on unlabeled data. The traditional semi-supervised algorithms perform the consistent regularization between teacher and student only at the last layer, which ignores the rich information learned in the intermediate layers from the large amount of unlabeled data. In contrast, our proposed semi-supervised method can learn efficient and rich multi-scale information from unlabeled RS images with our MSUC module. (2) Secondly, to solve the problem of high inter-class similarities, the key is to improve the feature extraction ability of the model. Therefore, we propose a Cross-Teacher-Student Attention module (CTSA) which makes the teacher network to help the student reconstruct the outputs of the encoder. Specifically, it takes the encoder result of the student as the query and the result of the teacher as the key and value. The benefit of such cross-network attention is that deep features can be learned from both the student and teacher models. It can help segment the hard-to-distinguish categories in RS images. For instance, as shown in Fig. \ref{fg1}, our semi-supervised model Fig. \ref{fg1} (h) can achieve more accurate segmentation results than the SOTA Unimatch method Fig. \ref{fg1} (g) on RS images. 

 
 Notably, previous studies have partially addressed the scientific challenges of rich multi-scale information and high inter-class similarities in RS fully supervised semantic segmentation models‌\cite{cai2024poly,liu2024desformer,wang2023multiscale,zhao2021semantic}. However, in the field of semi-supervised RS, our paper is the first attempt to simultaneously address both the scientific challenges and provide an open-source implementation. We first time propose the MUCA model from the viewpoint of computing multi-scale uncertainty consistency regularization and using Cross-Teacher-Student attention. It prioritizes architectural innovations for RS semi-supervised semantic segmentation, advancing semantic segmentation performance under limited data. The proposed model is non-intrusive and can be easily integrated into existing semantic segmentation models without changing the network structure of the model itself. Overall, the major contributions of this work can be summarized as follows:

\begin{enumerate}
  \item We propose a novel semi-supervised model MUCA, specifically, to address the challenge of rich multi-scale information and  high inter-class similarities in RS semantic segmentation.
  \item To learn rich multi-scale information, MUCA introduces multi-scale uncertainty consistency regularization that constrains the consistency among feature maps at  different layers of the network. 
  \item To distinguish the high inter-class similarities, MUCA utilizes a new cross-teacher-student attention to guide the student network in reconstructing discriminative encodings by the teacher.
  \item Extensive experimental results demonstrate the efficacy of our MUCA model on the LoveDA and ISPRS-Potsdam datasets, compared to some SOTA methods. The results show promising achievement in RS semi-supervised segmentation tasks and highlight the significance of our contributions.
\end{enumerate}
The rest of this article is organized as follows: Section \ref{RW} briefly introduces the related work. In Section \ref{METHODS}, our model MUCA are proposed and discussed. Section \ref{Experiment} shows the experimental results and compares them with other SOTA methods. Section \ref{Conclusion} concludes the article.

\section{Related work}
\label{RW}
In recent years, the developments of deep neural networks have encouraged the emergence of a series of works on semi-supervised semantic segmentation of RS images. This section gives a short description of recent developments in related fields.
\subsection{RS semantic segmentation}
Semantic segmentation is to segment an image into regional blocks with certain semantic meanings \cite{sun2021gaussian,shi2025rethinking}, and recognize the specific semantic category of each regional block respectively. It commonly labels the image pixel by pixel \cite{zhou2023adaptive, xiong2024earthnets, li2024vision,hong2023cross,qiao2025sam}. Early semantic segmentation models including FCN \cite{deng2023dual}, SegNet \cite{zhou2022eca}, and U-Net \cite{wu2022uiu}, achieved remarkable performance in traditional semantic segmentation tasks. With the success of Transformers in vision in recent years, more and more semantic segmentation models are adopting the Transformers architecture\cite{wang2023hunting}. Complex backgrounds as well as resolution variations are the main challenges in semantic segmentation of RS images. In RS images, large intra-class and small inter-class variations between objects make image feature characterization difficult. Workman et al. \cite{workman2023handling} argued that samples with high-resolution labels can be used to guide the training process in supervised learning using low-resolution labels. Region aggregation methods have also been used to improve the resolution of images. Quan et al. \cite{quan2022building} utilized multi-scale edge features obtained by Differential Difference of Gaussian (DoG) methods to improve the results of edge extraction. Li et al. \cite{li2023progressive} proposed a progressive recurrent neural network to remove RS images destriping. Zhong et al. \cite{zhong2022nt} proposed a transformer-based noise identification network model to help mitigate the over-segmentation phenomenon. However, semantic segmentation algorithms rely heavily on a large amount of training data, and it is difficult for these frameworks to achieve good results when labeled data is limited.

\begin{figure*}[!t]
\centering
\includegraphics[width=7in]{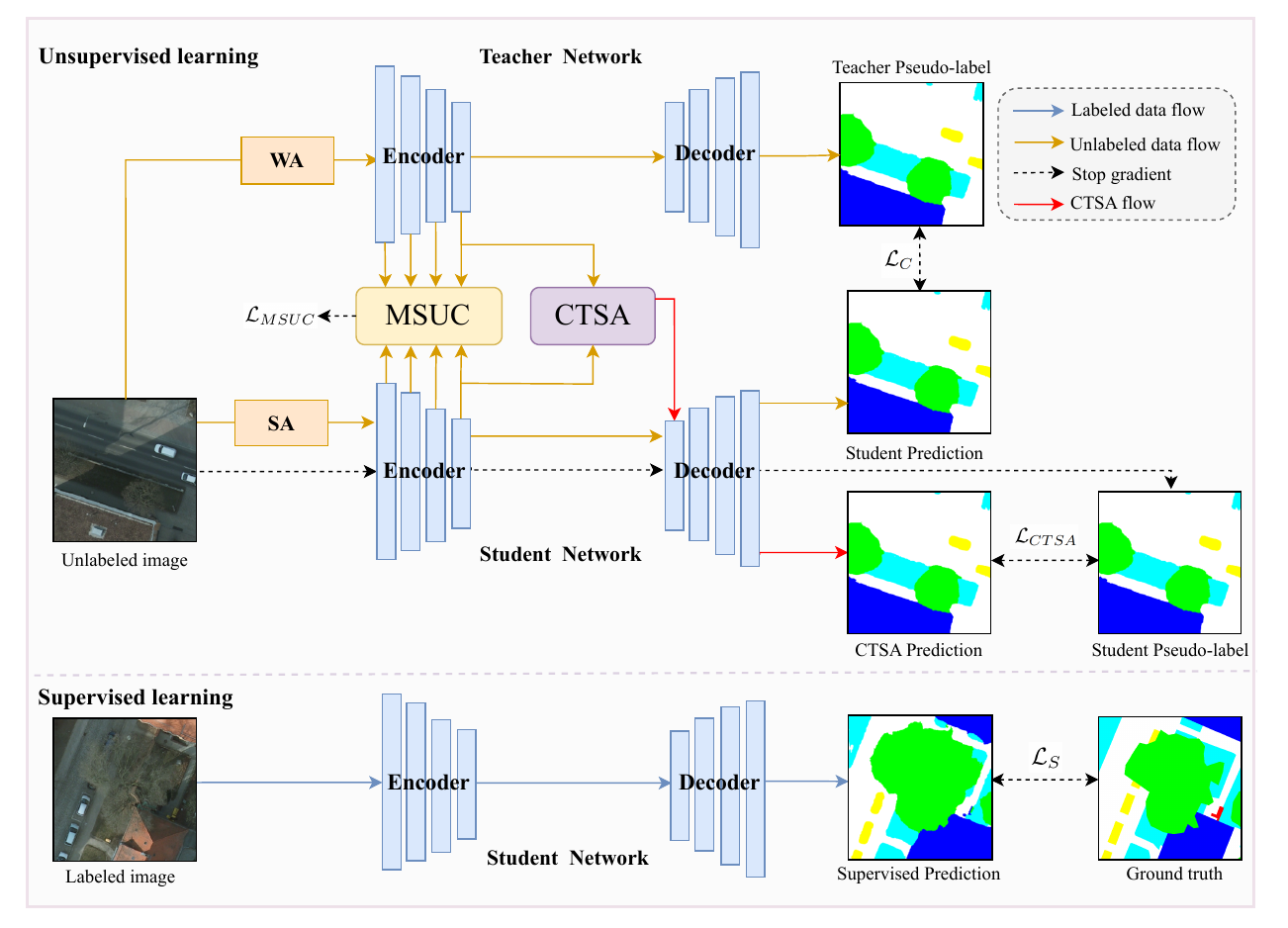}\vspace{-0.5cm}
\caption{Overall structure of our semi-supervised semantic segmentation model for RS images. The top and bottom sections correspond to the unsupervised learning for unlabeled data and ‌supervised learning for labeled data, respectively. Blue lines indicate labeled data flow, yellow lines indicate unlabeled data flow, red lines indicate CTSA flow, and dashed lines indicate data flow where gradient updating stops. The teacher network and student network interaction part contains the MSUC module and CTSA module proposed in this paper.}
\label{framework}
\vspace{-0.5cm}

\end{figure*}

\subsection{Semi-supervised Semantic Segmentation}
Semi-supervised semantic segmentation algorithms focus on how to better utilize large amounts of unlabeled data than supervised algorithms. Most semi-supervised semantic segmentation models use a basic convolutional neural network as the backbone, and implement semi-supervised algorithms in three different strategies.

The first strategy is pseudo-labeling for self-training, which involves generating pseudo-labels for unlabeled images based on a previously trained model on labeled data. It trains the model with newly generated images with pseudo-labels \cite{du2022learning, ke2022three, yuan2021simple}. The second strategy is to optimize the model by optimizing the consistent regularized loss function of feature perturbations \cite{jin2022semi, wang2023hunting, wang2022semi, olsson2021classmix}. Such methods utilize data-enhancement techniques to apply perturbations directly to the input image. They force the model to predict the same labels for both original and enhanced images. Some feature-based perturbation methods add internal perturbations to the segmentation network, resulting in a modified feature \cite{ouali2020semi}. UniMatch\cite{yang2023revisiting} took into account the nature of the semantic segmentation task and merged appropriate data augmentation into FixMatch \cite{sohn2020fixmatch}, which has evolved into a concise baseline of semi-supervised semantic segmentation algorithms.
The last strategy is teacher-student consistency \cite{chen2021semi}. The most classical method based on consistency regularization is the Mean-teacher \cite{tarvainen2017mean}. It considers the consistency between the predictions of the student and the teacher network. The weights of the teacher network are calculated from the exponential moving average (EMA) of the weights of the student's network. Recently, a few works have applied the teacher-student consistency method to the field of semi-supervision of RS images \cite{huang2024decouple}.

\subsection{Semi-Supervised Semantic Segmentation in RS Images}
Semi-supervised RS semantic segmentation has drawn more and more attention to make better use of the large amount of unlabeled RS image data \cite{fang2020collaborative, wang2023stcrnet, cai2024consistency, xuan2024tsg}. These RS semi-supervised semantic segmentation models can be broadly categorized into two groups.

The first one is to address limitations of semi-supervised models. For instance, Lu et al. \cite{lu2022simple} introduced a streamlined semi-supervised framework employing linear sampling to dynamically assign class thresholds, thereby enhancing pseudo-label quality. Li et al. \cite{li2021semisupervised} devised a semi-supervised strategy under a GAN framework. It optimizing model performance through consistency self-training that learns joint distributions of labeled and unlabeled data. This method leverages pixel-level training labels to assess prediction confidence. Ma et al. \cite{ma2023diversenet} critically observed that existing consistency learning frameworks based on network perturbations suffer from excessive complexity. To address this, they developed a divergence network exploring lightweight multi-head and multi-model perturbations to simultaneously enhance feature diversity and pseudo-label precision. However, these frameworks primarily focus on semi-supervised algorithmic shortcomings without considering the unique characteristics of RS images.

The second category focuses on enhancements targeting domain-specific challenges in RS. For instance, Gao et al. \cite{gao2024mcmcnet} developed a semi-supervised road extraction network via multiple consistency and multitask constraints (MCMCNet). It ensures enhanced preservation of road continuity and trunk integrity. Chen et al. \cite{chen2023semiroadexnet} proposed an adversarial learning-based semi-supervised network containing dual discriminators for road extraction from RS images. The two discriminators reinforce feature distribution alignment while maintaining consistency between labeled and unlabeled data. However, these semi-supervised methods typically focus on particular remote classes such as roads and traffic signs, demonstrating limited generalization. For the more widely used multi-class RS images, Xin et al. \cite{xin2024confidence} proposed a hybrid framework integrating consistency regularization and contrastive learning to address interclass similarities and intraclass variations in RS imagery. Nevertheless, their experiments predominantly focused on a 25\% labeled data ratio, failing to demonstrate effectiveness under very few labeled data scenarios. Liu et al. \cite{liu2023multi} propose a Multi-Level Label-Aware (MLLA) semi-supervised framework for remote sensing scene classification to address inter-class similarities. By extending semantic information extraction from unlabeled data from single-level to multi-level, MLLA enhances scene classification performance. Zhang et al.\cite{zhang2023mdmasnet} propose the MDMASNet framework to tackle the rich multi-scale information in remote sensing images.‌ ‌MDMASNet employs a pyramid architecture that integrates deformable convolution and dilated convolution to extract features from objects of diverse shapes and sizes. Although these networks target rich multi-scale information and high inter-class similarities in semi-supervised remote sensing, their proposed‌ network models ‌exhibit excessive complexity and fail to achieve non-intrusiveness. Huang et al.\cite{huang2023semi} considered the significant domain shift between different RS datasets. They proposed the semi-supervised bidirectional alignment for RS cross-domain scene classification. To address the inevitable issue of erroneous pseudo-labels and long-tail distribution in RS images, Huang et al.\cite{huang2024decouple} introduced a Decoupled Weighting Learning (DWL) framework. During training, the decoupled learning module separates the predictions of labeled and unlabeled data. And the ranking weighting module attempts to adaptively weight each pseudo-label of the unlabeled data based on its relative confidence rank within the pseudo-class.

However, these RS semi-supervised models do not simultaneously address the rich multi-scale information and high inter-class similarities in RS images. Additionally, some RS semi-supervised models do not provide open-source code, limiting their utility for subsequent work. This paper is the first attempt to simultaneously address both challenges in RS semi-supervised field and provide an open-source implementation. We proposes the MUCA semi-supervised model to tackle these challenges and improve the performance of RS semi-supervised semantic segmentation.

\section{METHODS}
\label{METHODS}
This section is organized as follows: Section \ref{Objectives} describes the main optimization objectives, Section \ref{mu} introduces the principles of the multi-scale uncertainty consistency module, and Section \ref{ctsa} presents the basic principles of the cross-teacher-student attention module. The overall structure of our approach is shown in Fig. \ref{framework}. 

\subsection{Main Optimization Objectives}
\label{Objectives}
We define $ D^L=\{(x_i^l,y_i)\}_{i=1}^{N_L} $ as labeled data and $ D^U=\{(x_i^u)\}_{i=1}^{N_U} $ as unlabeled. Here ${x_i^l}\in \mathbb{R}^{H\times W \times 3}$ denotes the labeled image, ${y_i}\in \mathbb{R}^{H\times W \times K}$ is the ground truth of $K$ classes, while ${x_i^u}\in \mathbb{R}^{H\times W \times 3}$ denotes the unlabeled image, $N_L$ and $N_U$ are the amount of labeled and unlabeled images. $H$ and $W$ specify the height and width of the image. It is worth noting that in general $N_U$ is much larger than $N_L$. For unlabeled images, weak augmentation (WA) and strong augmentation (SA) are performed to train the teacher and student networks respectively. The main loss function is:
\begin{equation}
\label{deqn_ex1a}
\mathcal L=\mathcal L_S+\mathcal L_U =\frac{1}{N_L}{\sum_{i=0}^{N_L}\mathcal L_{CE}(p_i^l,y_i)}+ \mathcal L_U,
\end{equation}
\noindent where $\mathcal L_S$ represents the loss of supervised learning with labeled data, and $\mathcal L_U$ represents the loss with unlabeled data, $p_i^l$ is the prediction result of the labeled image $x_i^l$, $\mathcal L_{CE}$ is the cross-entropy loss. More specifically, for $\mathcal L_U$, it consists of three parts.
\begin{equation}
\label{deqn_ex1a}
\mathcal L_U =\mathcal L_{C}+\mathcal L_{MSUC}+\mathcal L_{CTSA}.
\end{equation}
\noindent The first part $\mathcal L_C$ is the basic teacher-student consistency loss function, similar to the Mean-teacher model\cite{pelaez2023survey}. The second part is the loss function $\mathcal L_{MSUC}$ for multi-scale uncertainty consistency, and the third part $\mathcal L_{CTSA}$ is related to cross-teacher-student attention, which will be introduced later.
\subsection{Multi-Scale Uncertainty Consistency Module}
\label{mu}
\begin{figure}[h]
\centering
\includegraphics[width=3.7in]{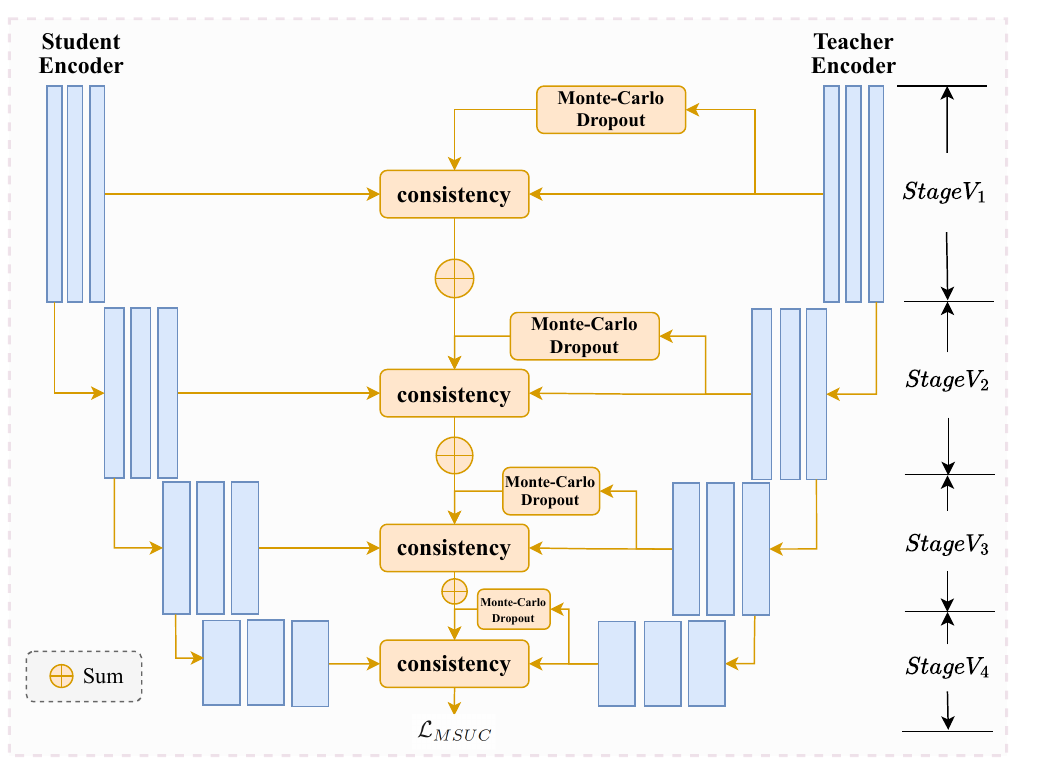}
\caption{MSUC model structure. Uncertainty is computed for the feature layer at each stage of the teacher and student networks, and features with high uncertainty are discarded when computing consistency regularization.}
\label{MSUC}\vspace{-0.5cm}
\end{figure}

Basic models of teacher-student architectures update the teacher's weights $\theta^t$ by the exponential moving average (EMA) with the student's weights $\theta^s$. It integrates the information from different training steps. The teacher's weight $\theta_t^t$ is updated at training step $t$ as $\theta_t^t= \alpha\theta_{t-1}^t + (1-\alpha)\theta_{t}^s$, where $\alpha$ is the EMA decay. This basic teacher-student architecture is widely used in semi-supervised networks. However, it is critical to take into account the inherent difference between RS and natural images. Objects in RS images exhibit a wide range of scales, spanning from large buildings to small vehicles, and the same object can appear at varying sizes under different resolutions. Therefore, semi-supervised learning in RS images is difficult to extract effective features. To solve this problem, we propose a Multi-Scale Uncertainty Consistency (MSUC) module for the student to progressively learn multi-scale features from reliable targets, as shown in Fig. \ref{MSUC}. Given a batch of training images, the teacher model not only generates pseudo-labels for target prediction, but also estimates the uncertainty of each target at multiple feature levels. It optimizes the student model through multilevel consistency loss. This will motivate the model to learn different layers and sizes of features from RS images.

\subsubsection*{\bf Uncertainty Estimation}
We first compute pseudo-label uncertainty to mitigate the impact of low quality pseudo-label in semi-supervised learning frameworks. This is to enable the model to progressively learn more reliable targets from multi-scale information. Specifically, we use Monte-Carlo Dropout\cite{gal2016dropout} to estimate uncertainty. Monte-Carlo method can not only calculate the uncertainty between different classes but also evaluate the uncertainty in the multiple predictions of a neural network. The basic principle of Monte-Carlo Dropout is to make predictions of $T$ times for the same sample with the same model. These $T$ predictions are different, and their variance is calculated to compute the model uncertainty. In detail, we perform $T$ random forward passes for the teacher model, each of which uses the random dropout and noise. Thus, for each pixel in the image, we obtain a set of softmax probability vectors: $\{p_t\}_{t=1}^T$. We choose the prediction entropy as a measure of the approximate uncertainty because it has a fixed range. Formally, the prediction entropy can be summarized as:
\begin{equation}
\label{u_k}
u_k =\frac{1}{T}\sum_{t=1}^{T}p_t^k,
\end{equation}
\begin{equation}
\label{u}
u =-\sum_{k=1}^{K}u_k{log{(u_k)}},
\end{equation}
\noindent where $p_t^k$ is the result of pixels of the class $k$ in the $t^{th}$ prediction procedure. In this way, the uncertainty of multiple predictions for the same class can be calculated. 
According to Eq. \ref{u_k} and Eq. \ref{u}, the value $u$ increases when the network provides completely opposite predictions many times. Moreover, Eq. \ref{u_k} and Eq. \ref{u} can also take into account the uncertainty between different classes. If the neural network gives similar predictions across multiple classes, the uncertainty value $u$ will also increase.

\subsubsection*{\bf Consistency Loss Functions for Multiscale Uncertainty}
To formally describe MSUC, we denote the encoded features from the four stages of the encoder as $V_i \in {\mathbb{R}^{C_i×H_i×W_i}}$, $i \in \{1,2,3,4\}$, where $C_i$, $H_i$, and $W_i$ denote the number of channels, the height, and the width of the feature maps from the $i^{th}$ stage, as shown in Fig. \ref{MSUC}. More specifically, $V_1$ is the output of the first stage with the lowest abstraction but the highest spatial resolution, and $V_4$ is the output of the fourth stage with the highest abstraction but the lowest spatial resolution. The final consistency loss function was computed by aligning the encoded visual features at each stage of the student model and the teacher model.
\begin{equation}
\label{l_msuc}
\mathcal L_{MSUC}= \sum_{i=1}^4{\frac{\sum_m\mathbb{I}(u_{i_m}<H)\mathcal L_{\varrho}(V^t_{i_m}-V^s_{i_m})}{\sum_m\mathbb{I}(u_{i_m}<H)}},
\end{equation}
\noindent where $V_{i_m}^t$ and $V_{i_s}^s$ are the outputs of the teacher and student encoders at the pixel $m$ of the $i^{th}$ stage, $u_{i_m}$ is the uncertainty at the pixel $m$ of the $i^{th}$ stage, $\mathbb{I}$ is the indicator function, and $H$ is a threshold. Specifically, for threshold setting, we adopt a dynamic threshold approach using a Gaussian ramp-up curve\cite{laine2022temporal}. During the training process, the uncertainty threshold $H$ is gradually increased from the initial $\frac{1}{2}H_{max}$ to $H_{max}$, where $H_{max}$ denotes the maximum uncertainty threshold setted as $ln2$ in our experiments. Consequently, as training progresses, our method filters out progressively fewer samples, enabling the student model to gradually learn from relatively certain to more uncertain cases. This design ensures model stability in early training phases by eliminating substantial low-quality pseudo-label pixels. As the training progresses, the model becomes gradually more reliable, allowing more samples to be selected in the later stages of training, even if their uncertainty is slightly higher. Our $H_{max}$ setting still filters out pseudo-label pixels with higher uncertainty. $\mathcal L_{\varrho}$ is the Huber loss function, which is formulated as follows:
\begin{equation}
\label{deqn_ex1a}
\mathcal L_{\varrho}(V^t_{i_m}-V^s_{i_m})= \begin{cases}
\frac{1}{2}(V^t_{i_m}-V^s_{i_m})^2,  & \text{if}|V^t_{i_m}-V^s_{i_m}| \leq \varrho, \\
 {\varrho} |V^t_{i_m}-V^s_{i_m}|-\frac{1}{2}{\varrho}^2 ,
        &\text{otherwise},
\end{cases}
\end{equation}
\noindent where $\varrho$ is the soft threshold for Huber loss, which is set to 1.0 in this paper.
\subsection{Cross-Teacher-Student Attention}
\label{ctsa}
\begin{figure}[!t]
\centering
\includegraphics[width=3.5in]{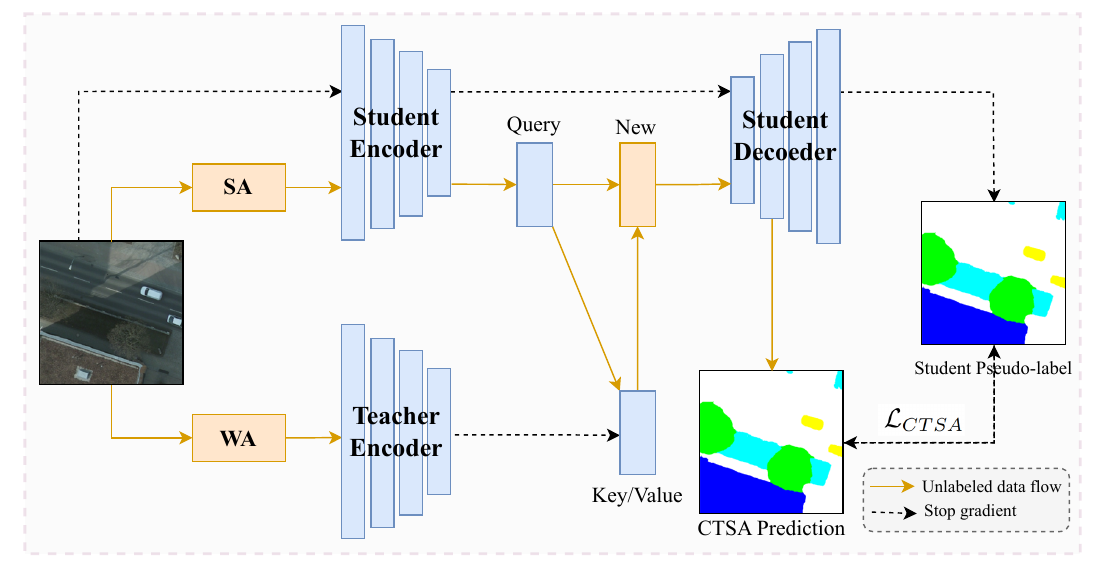}
\caption{CTSA model structure. An attention approach allows the teacher network to guide the student network in reconstructing the student network's encoder results.}
\label{CTSA}\vspace{-0.5cm}
\end{figure}
We propose the Cross-Teacher-Student Attention (CTSA) model, as illustrated in Fig. \ref{CTSA}, to promote the ability of semi-supervised segmentation method for objects with high inter-class similarities.

Consistency regularization methods are based on the assumption of smoothness\cite{chapelle2009semi}. This means that a robust model should produce similar predictions for points and their variants with noise. In other words, a model trained with consistency methods should not be affected by different perturbations added to the data. However such semi-supervised frameworks ignore the inherent domain gap between natural and RS images. One RS image contains much richer content than one natural image, and objects of different categories are commonly intertwined. Therefore perturbations will make the RS image contain more confusing information. The features learned by the neural network from different perturbed RS images may lead to some outliers as pseudo-annotations, misleading the training of the model on unannotated data. 

To address this problem, we use the features extracted by the student and teacher networks via different perturbations to reconstruct the features of the unlabeled data. We use the encoder result of the student as $query$ and the encoder result of the teacher as $key$ and $value$. The teacher guides the reconstruction of the encoder of the student, as shown in Fig. \ref{CTSA}. Specifically, we calculate the similarity between each channel of the unlabeled features of the student and teacher networks. The channels with higher similarity are more important in reconstructing the unlabeled features. This ensures enhanced feature extraction within the same category while reducing cross-category feature weights. 

More importantly, the new features constructed through CTSA effectively integrate the advantages of the teacher and student network encoders. Specifically, the SA data fed into the student encoder provides the original features which are diverse and rich. Meanwhile, the teacher encoder, receiving WA data, reconstructs another set of smooth and stable features. The decoder is trained using features generated from strongly supervised data along with features reconstructed by CTSA, resulting in better optimization outcomes. Particularly in the early stages of semi-supervised learning, where training instability and low-quality pseudo-labels often hinder efficient convergence, the teacher network encoder provides more easily learnable features for the student network decoder. This improves the model's stability, mitigates training fluctuations, accelerates convergence, and enhances training efficiency.

We further clarify the CTSA formulation as follows. Given the student and teacher characteristics $V_4^s$ and $V_4^t$ after the fourth stage of the encoder, we set $V_4^s$ as $query$, $V_4^t$ as $key$ and $value$, in a multi-head manner:
\begin{equation}
\label{deqn_ex1a}
F^{V_4^s}=flatten(V_4^s),\ F^{V_4^t}=flatten(V_4^t),
\end{equation}
\begin{equation}
\label{deqn_ex1a}
q=F^{V_4^s}w_q,\ k=F^{V_4^t}w_k,\ v=F^{V_4^t}w_v,
\end{equation}
\noindent where $w_q,w_k,w_v\in\mathbb{R}^{C\times 2C}$ are the transformer weights, $F^{V_4^s},F^{V_4^t}\in \mathbb{R}^{C\times d}$, $d$ is the number of patches and $C$ is the channel dimension. The CTSA is defined as:
\begin{equation}\label{deqn_ex1a}
V_{out}=\phi[\psi(q^\mathrm{ T } k)v^{ \mathrm{ T }}w_{out}],
\end{equation}
\noindent where $\psi$ denotes instance normalization and $\phi$ is softmax function. $V_{out}$ is fed into the decoder to get the final predicted value of the CTSA module as described below:
\begin{equation}
\label{deqn_ex1a}
p^u=Decoder(V_{out}).
\end{equation}
The loss function for the CTSA module is calculated as Eq. \ref{eq_ctsa}, where $\hat y_{i}$ is the generated pseudo-label from unlabeled data.
\begin{equation}
\label{eq_ctsa}
\mathcal L_{CTSA}=\frac{1}{N_U}{\sum_{i=0}^{N_U}\mathcal L_{CE}(p_i^u,\hat y_{i})},
\end{equation}

\begin{algorithm}[H]
\caption{Training process of MUCA}\label{alg:alg1}
\begin{algorithmic}
\STATE 
\STATE $ \textbf{Input:} $
\STATE \hspace{0.5cm}$   D_U^e=\{(x_i^u)\}_{i=1}^{N_U^e} $, $D_L^e=\{(x_i^l,y_i)\}_{i=1}^{N_L^e} $
\STATE$\textbf{Outpt}$: 
\STATE\hspace{0.5cm}$\Theta$: optimal model parameters
\STATE 1:\textbf{while} until converge:
\STATE 2:\hspace{0.75cm}\textbf{for} $x_i^l$ , $x_i^u$  in  $ D_L^e$ , $ D_U^e$:
\STATE 3:\hspace{1.25cm}$\mathcal L_{S}$$=CE(model\_s(x_i^l),y_i)$

\STATE 6:\hspace{1.25cm}$W\_x_i^u=WeakAugment(x_i^u)$
\STATE 7:\hspace{1.25cm}$S\_x_i^u=StrongAugment(x_i^u)$

\STATE 8:\hspace{1.25cm}$V^s,Enc_{s},Pre_{s}=model\_s(S\_x_i^u)$
\STATE 9:\hspace{1.25cm}$V^t,Enc_{t},Pseudo_{t}=model\_t(W\_x_i^u)$
\STATE 10:\hspace{1.15cm}$\mathcal L_{C}$$=Huber(Pre_{s},Pseudo_{t})$
\STATE 11:\hspace{1.15cm}\textbf{for} $j$ in [1:4]:
\STATE 12:\hspace{1.5cm}Calculate uncertainty $u$ via Eq. \ref{u} 
\STATE 13:\hspace{1.55cm}\textbf{if} $u<H$:
\STATE 14:\hspace{1.8cm} Calculate $\mathcal L_{MSUC}$ of $j$ stage via Eq. \ref{l_msuc}
\STATE 15:\hspace{1.8cm} $\mathcal L_{MSUC} +=\mathcal L_{MSUC}$
\STATE 16:\hspace{1.15cm}\textbf{end for} 
\STATE 17:\hspace{1.15cm}$Pseudo_{s}=model\_s(x_i^u).detach()$
\STATE 18:\hspace{1.15cm}$Pre_{ctsa}=model\_s\_d(CTSA(Enc_{s},Enc_{t}))$
\STATE 19:\hspace{1.15cm}$\mathcal L_{CTSA}$$=CE(Pre_{ctsa},Pseudo_{s})$
\STATE 20:\hspace{1.15cm}$\mathcal L_{U}=\mathcal L_{C}+\mathcal L_{CTSA}+\mathcal L_{MSUC}$ 
\STATE 21:\hspace{0.65cm}\textbf{end for} 

\STATE 22:\hspace{0.65cm}$\mathcal L=\mathcal L_{S}+\mathcal L_{U}$
\STATE 23:\hspace{0.65cm}Back propagating the $\mathcal L$ and update $\Theta$
\STATE 24:\hspace{0.65cm}Mark the current model checkpoint and use it to 
\STATE\hspace{1.10cm}predict pseudo labels of unlabeled images
\STATE 25:\hspace{0.65cm}Save the best checkpoint $\Theta$
\STATE 26:\textbf{return} $\Theta$
\end{algorithmic}
\label{alg1}
\end{algorithm}
Algorithm \ref{alg1} gives the core pseudo-code for the training process of MUCA on the unlabeled data. $D_U^e$ and $D_L^e$ are the unsupervised and supervised datasets for one epoch. $N_U^e$ and $N_L^e$ are number of images in $D_U^e$ and $D_L^e$. $model\_s $ and $model\_t$ denote the student and teacher models. $model\_s\_d$ is the decoder of the student model. $V^s$ and $V^t$ are the feature maps extracted by the encoders of the student network and the teacher network. $Enc_{s}$ and $Enc_{t}$ are the output results of the student network and teacher network encoders. $Pre_{s}$ represents the final prediction of the student network.  $Pseudo_{t}$ represents the pseudo-labels generated by the teacher network, and $Pseudo_{s}$ represents the pseudo-labels generated by the student network. $CE$ is the Cross-Entropy loss function, and $Huber$ is the Huber loss function.

\section{Experiment}
\label{Experiment}

In this section, we conduct experiments on semi-supervised semantic segmentation of RS images to evaluate the proposed MUCA model. We first perform ablation studies on MUCA to validate the effectiveness of the proposed modules. Then, we conducted comparative experiments to compare MUCA with several SOTA models. Our code is released at \href{https://github.com/wangshanwen001/RS-MUCA}{https://github.com/wangshanwen001/RS-MUCA}.

\subsection{RS Dataset}

\subsubsection*{\bf LoveDA}
The LoveDA RS dataset \cite{Loveda} consists of 5987 images and 166,768 annotated objects from three different cities. Each image is $1024 \times 1024$ with a spatial resolution of 0.3 meters. The dataset has seven segmentable classes, such as Building, Road, Water, Barren, Forest, Agriculture, and Background. Due to the memory limitations of the GPU, each image in the LoveDA dataset is resized and cropped to $512 \times 512$, producing a total of 16,764 cropped images for subsequent deep learning training. During training, we divide the dataset into training set, validation set, and test set by 6:2:2. 

\subsubsection*{\bf ISPRS-Potsdam}
The ISPRS-Potsdam RS dataset is provided to facilitate research on RS images semantic segmentation \cite{ISPRS_Potsdam}. This dataset has a resolution of 0.05 meters and consists of 38 super-large satellite RS images of $6000\times6000$ size. There are six segmentable classes in the dataset including Impervious surfaces, Building, Low vegetation, Tree, Car, and Background. To facilitate the training procedure, we crop the original image into $512 \times 512$ and the number of cropped images is 5472. The dataset is divided into training, validation and test sets by 6:2:2.

\subsection{Data Augmentation and Experiment Details}
We first augmentation the labeled images via geometric transformations (i.e., image scaling, horizontal flipping, vertical flipping and length and width warping) and Gaussian blurring. The weak augmentation used for unlabeled images is geometric augmentation, and strong augmentation is done by methods such as CutMix \cite{yun2019cutmix}. 

The experiments are conducted on A6000 (48G) with Python-3.8.10, Pytorch-1.13.0, and Cuda v11.7. The optimization algorithm is stochastic gradient descent with an initial learning rate 0.007. The learning rate is updated on each epoch, the weight decay is set to 0.0001, and the minimum learning rate is set to 0.00007. Then, we train the semi-supervised segmentation model using $1\%$, $5\%$ and $10\%$ of the labeled images and the remaining unlabeled images.

\subsection{ Evaluation metrics}
Most of the popular semantic segmentation metrics are employed to comprehensively evaluate the performance, e.g., intersection and concurrency ratio (IoU), F1-Score (F1), and Cohen's Kappa as show below:

\begin{equation}
\label{deqn_ex1a}
IoU=\frac{TP}{TP+FN+FP},
\end{equation}

\begin{equation}
\label{deqn_ex1a}
Recall=\frac{TP}{TP+FN},
\end{equation}

\begin{equation}
\label{deqn_ex1a}
Precision=\frac{TP}{TP+FP},
\end{equation}

\begin{equation}
\label{deqn_ex1a}
F1=\frac{2 \times Recall \times Precision}{Recall+Precision},
\end{equation}

\begin{equation}
\label{deqn_ex1a}
OA=\frac{TP+TN}{TP+TN+FP+FN},
\end{equation}

\begin{equation}
\label{deqn_ex1a}
PRE=\frac{(TP+FN)(TP+FP)+(TN+FN)(TN+FP)}{(TP+TN+FP+FN)^2},
\end{equation}
\begin{equation}
\label{deqn_ex1a}
Kappa=\frac{OA-PRE}{1-PRE},
\end{equation}
\noindent where $TP$ is the number of correctly predicted positive pixels, $TN$ is the number of correctly predicted negative pixels, $FP$ is the number of incorrectly predicted positive pixels, and $FN$ is the number of incorrectly predicted negative pixels. In our experiments we average $IoU$ and $F1$ across all classes, so the final experimental metrics also include mean $IoU$ ($mIoU$) and mean $F1$ ($mF1$). Then, $mIoU$ and $mF1$ is calculated as follows:
\begin{equation}
\label{deqn_ex1a}
mIoU=\frac{\sum_{i=1}^K{IoU_i}}{K},
\end{equation}

\begin{equation}
\label{deqn_ex1a}
mF1=\frac{\sum_{i=1}^K{F1_i}}{K},
\end{equation}
\noindent where $IoU_i$ means the IoU of the $i^{th}$ class and $K$ is the number of classes.
\subsection{Ablation study}
\subsubsection*{\bf Ablation of components}
We ablate each component step by step to investigate their performance. It is worth noting that we also evaluate the multi-scale consistency loss function of MSUC module without uncertainty, which is denoted as NoUC. Differing from the MSUC module, NoUC only uses standard consistency regularization to compute the loss for each stage feature map. The ablation studies are conducted on the ISPRS-Potsdam dataset with labeled data ratio of $5\%$. The baseline is the supervised model with $5\%$ labeled data. The semantic segmentation model used in this subsection is SegFormer-B2.

\begin{table}[h]
\caption{Ablation study on the effectiveness of components.\label{tab:table1}}
\centering
\begin{tabular}{c| c c c|c}
		\toprule
		\textbf{Dataset}&\textbf{NoUC} & \textbf{MSUC}  & \textbf{CTSA}&\textbf{mIoU ($5\%$)} \\
		\midrule
		 \multirow{5}{*}{ISPRS-Potsdam}~ & ~ & ~ &~ &  72.01 \\ 
          ~ &\checkmark & ~ & ~ & 73.14  \\
		 ~ & ~& \checkmark & ~  & 73.95 \\
		 ~ & ~& ~ & \checkmark  & 73.80 \\
		 ~ & ~& \checkmark & \checkmark  & 74.62\\
         \midrule
         \multirow{5}{*}{LoveDA}~ & ~ & ~ &~ &  44.20 \\ 
          ~ &\checkmark & ~ & ~ & 48.21 \\
		 ~ & ~& \checkmark & ~  & 50.02 \\
		 ~ & ~& ~ & \checkmark  & 49.81 \\
		 ~ & ~& \checkmark & \checkmark  & 50.97\\
		\bottomrule
	\end{tabular}
\end{table}

NoUC, which despite solely utilizes standard consistency regularization to compute the loss function at each stage, still improves semantic segmentation performance with limited labeled data. Compared with NoUC, MSUC achieves improved mIoU on both the ISPRS-Potsdam dataset and LoveDA dataset. These results indicate that uncertainty computation enables the student network to discard unreliable information. ‌The model integrated with the CTSA module demonstrates a significant mIoU improvement over the baseline. This demonstrates that the CTSA module effectively enhances the segmentation performance of semi-supervised model on RS imagery.

\textbf{Should we integrate the CTSA module in the test stage?} 
The CTSA module requires both teacher-student networks to complete the attention. However, generally speaking, we only save one final checkpoint file. In this study, we save both the teacher and student checkpoint files in each epoch. We then investigate the influences of CTSA on the $mIoU$ in the testing stage. As shown in Table \ref{tab:table2}, the impact of the CTSA module is almost negligible in the testing phase. The double disk consumption and limited performance improvement make us abandon the CTSA module in the testing phase. 

\begin{table}[h]
\caption{The impact of CTSA module in the test stage.\label{tab:table2}}
\centering
\begin{tabular}{ c|c |c |c}
		\toprule
		\textbf{Dataset}&\textbf{Stage}  & \textbf{CTSA}&\textbf{mIoU ($5\%$)} \\
		\midrule
		 \multirow{2}{*}{ISPRS-Potsdam} &\multirow{2}{*}{Test} & ~ & 73.80  \\ 
           ~ &~ & \checkmark & 73.87  \\
           \midrule
           \multirow{2}{*}{LoveDA} &\multirow{2}{*}{Test} & ~ & 49.81 \\ 
           ~ &~ & \checkmark & 49.85 \\
		\bottomrule
	\end{tabular}
\end{table}

\subsubsection*{\bf Analysis of multi-level feature maps of MSUC module}
MSUC encodes visual features from the four stages of the encoder, i.e., $V_i \in {\mathbb{R}^{C_i×H_i×W_i}}$, $i \in \{1,2,3,4\}$. Therefore, it is critical to explore the impact of multi-level (i.e., low to high) deep features on our MSUC. From Table \ref{tab:table3}, we can see that the deep MSUC feature $V_4$ achieves greater improvement than the other low-level features, while the lowest feature $V_1$ makes the worst mIoU. The reason is that when only utilizing features extracted from a single stage of the model, deep features are more effective than low-level features on forming global semantic representations such as object categories and overall contours.

\begin{table}[h]
\caption{Analysis of multi-level feature maps of MSUC module.\label{tab:table3}}
\centering
\begin{tabular}{c|c c c c |c}
		\toprule
		\textbf{Dataset}&\textbf{$V_1$}&\textbf{$V_2$} & \textbf{$V_3$}  & \textbf{$V_4$}&\textbf{mIoU ($5\%$)} \\
		\midrule
            \multirow{5}{*}{ISPRS-Potsdam}~&~&~ & ~ & ~ &  72.01 \\
		 ~&\checkmark&~ & ~ & ~ &  72.41 \\ 
          ~&~& \checkmark & ~ & ~ & 72.55  \\
		 ~&~&~ & \checkmark & ~  & 72.76\\
		 ~&~&~ & ~ & \checkmark  & 72.96\\
         \midrule
         \multirow{5}{*}{LoveDA}~&~&~ & ~ & ~ &  44.20 \\
		 ~&\checkmark&~ & ~ & ~ &   45.97 \\ 
          ~&~& \checkmark & ~ & ~ &  46.72  \\
		 ~&~&~ & \checkmark & ~  &  47.09 \\
		 ~&~&~ & ~ & \checkmark  &   47.91\\
		\bottomrule
	\end{tabular}
\end{table}

\subsubsection*{\bf Analysis of features fusion for MSUC module}
We fuse deep and low features stage by stage for the MSUC module. Specifically, we firstly employ the deep feature $V_4$ alone, and then gradually add all the other level features. Table \ref{tab:table4} shows the results of these different combinations. From the results, we observe that the mIoU gradually increases as lower level features are added stage by stage. This is because ‌low-level features‌, generated by the shallow stages of the network, have small receptive fields and primarily capture ‌local details‌ such as image edges and textures. Deep features‌ accumulate ‌large-scale contextual information‌ through repeated downsampling and convolutional operations, forming ‌global semantic representations‌ like object categories and overall contours. Therefore, the ‌high resolution‌ and ‌detail retention capability‌ of low-level features make them inherently suited for ‌fine-grained analysis‌. While the ‌lower resolution‌ and ‌broader receptive fields‌ of deep features enable them to excel in ‌large-scale semantic comprehension. The fusion of deep and low features improves the consistency of the semi-supervised model at different stages and scales, significantly improving the model's capability to capture multi-scale characteristics of target objects. ‌Consequently, it optimizes semantic segmentation performance in RS images.

\begin{table}[h]
\caption{Analysis of deep and low features fusion for MSUC module.\label{tab:table4}}
\centering
\begin{tabular}{c|c c c c |c}
		\toprule
		\textbf{Dataset}&\textbf{$V_1$}&\textbf{$V_2$} &  \textbf{$V_3$}  & \textbf{$V_4$}&\textbf{mIoU ($5\%$)} \\
		\midrule
       \multirow{5}{*}{ISPRS-Potsdam}& ~&~ & ~ & ~ &  72.01 \\ 
		 ~&~&~ & ~ & \checkmark &  72.96 \\ 
          ~&~& ~ & \checkmark & \checkmark &  73.42 \\
		 ~&~&\checkmark & \checkmark & \checkmark  & 73.66 \\
		 ~&\checkmark&\checkmark & \checkmark & \checkmark  & 73.95\\
         \midrule
         \multirow{5}{*}{LoveDA}& ~&~ & ~ & ~ &  44.20 \\ 
          ~&~&~ & ~ & \checkmark &  47.91 \\ 
          ~&~& ~ & \checkmark & \checkmark &  49.10  \\
		 ~&~&\checkmark & \checkmark & \checkmark  & 49.64  \\
		 ~&\checkmark&\checkmark & \checkmark & \checkmark  & 50.02 \\
		\bottomrule
	\end{tabular}
\end{table}

\subsubsection*{\bf Confusion Matrix Visualization}
\begin{figure}[h]
\centering
\includegraphics[width=3.5in]{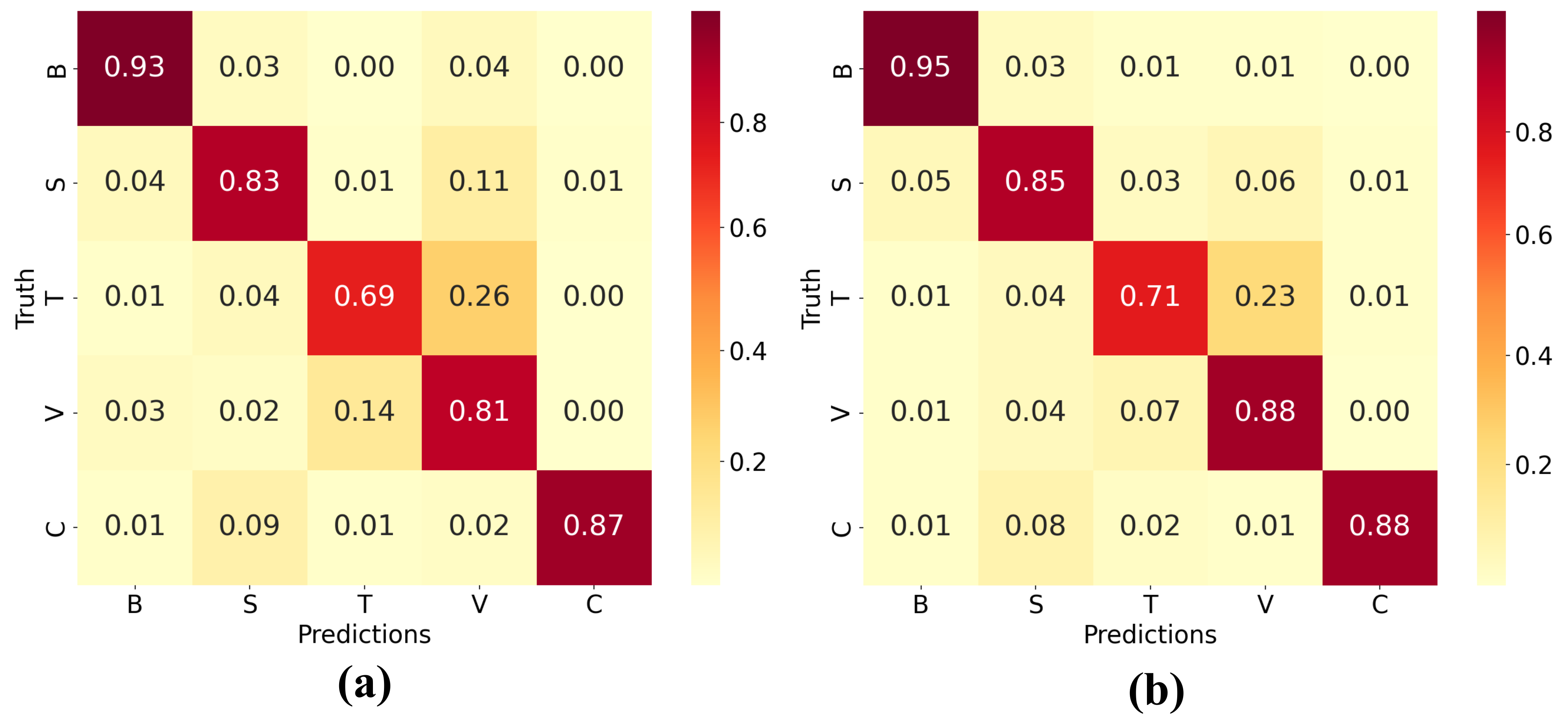}
\caption{Visualization confusion matrix of model without MUCA (a) and with MUCA (b) on ISPRS-Potsdam dataset. Car (C), Low vegetation (V), Tree (T), Impervious surfaces (S), and Building (B).}
\label{conf_Potsdam}
\end{figure}

\begin{figure}[h]
\centering
\includegraphics[width=3.5in]{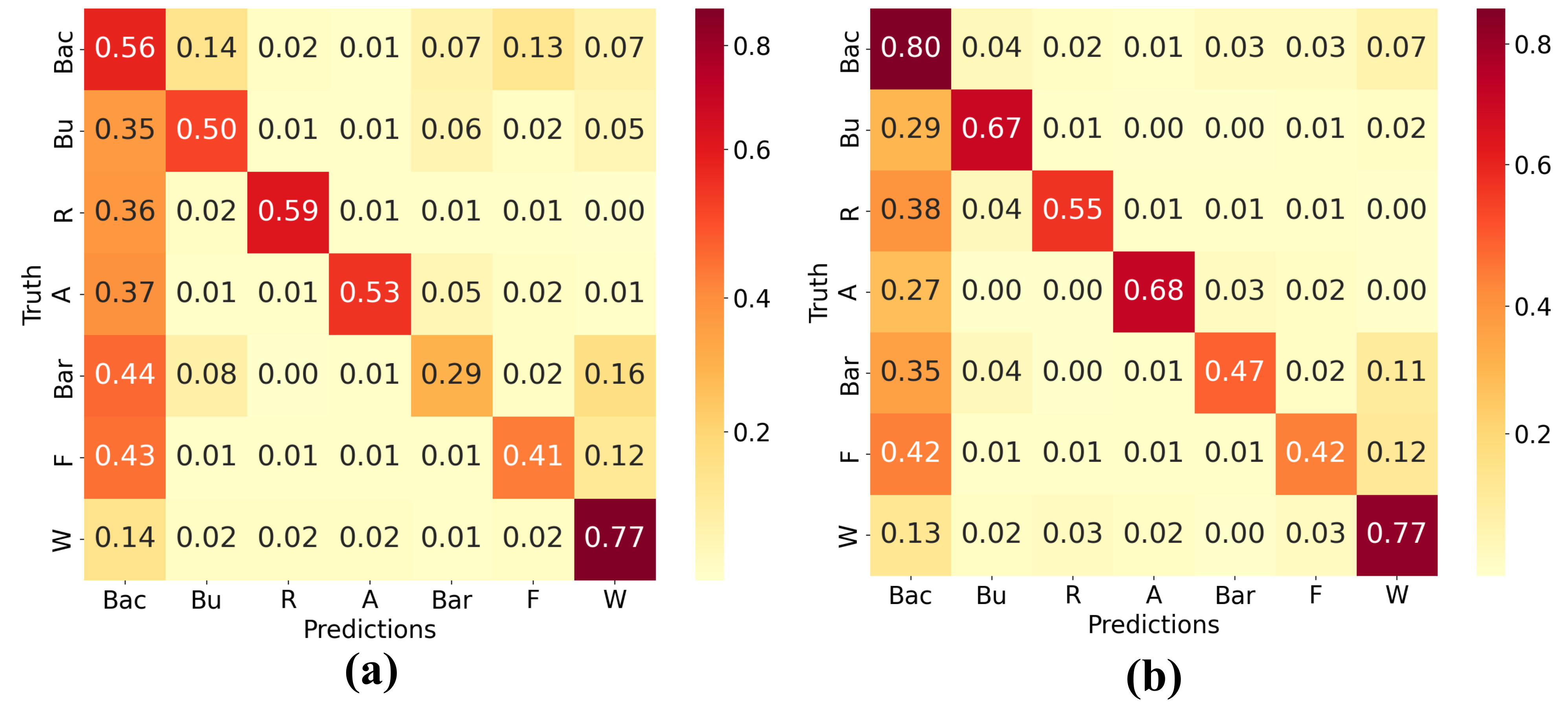}\vspace{-0.5cm}
\caption{Visualization confusion matrix of model without MUCA (a) and with MUCA (b) on LoveDA dataset. Background (Bac), Building (Bu), Road (R), Agriculture (A), Barren (Bar), Forest (F), and Water (W).}
\label{conf_LoveDA}

\end{figure}

We visualized the confusion matrices of the ISPRS-Potsdam dataset at labeled data ratio of 5\%, as shown in Fig. \ref{conf_Potsdam} and Fig. \ref{conf_LoveDA}. It intuitively demonstrates the improvement of the semi-supervised semantic segmentation with MUCA. We perform a row normalization on the confusion matrix, where the redder of the color the closer to the truth. We can see that the introduced MUCA significantly improves the ability to distinguish hard-to-differentiate classes such as Building, Impervious surfaces, Low vegetation and Tree on ISPRS-Potsdam dataset. Furthermore, it demonstrates significant enhancement in segmenting Barren, Agriculture, Building, and Background classes on the LoveDA dataset.

\vspace{-0.5cm}

\subsection{Comparison with SOTA methods on RS Datasets}
This section conducts experiments on ISPRS-Potsdam and LoveDA datasets, compared to the SOTA methods, including Mean teacher \cite{tarvainen2017mean}, CutMix \cite{yun2019cutmix}, CCT \cite{ouali2020semi}, CPS \cite{chen2021semi}, LSST \cite{lu2022simple}, Fixmatch \cite{sohn2020fixmatch}, Unimatch \cite{yang2023revisiting}, DWL \cite{huang2024decouple}, and Allspark \cite{Wang_2024_CVPR}. Specifically, we show the results for labeled data ratios of $1\%$, $5\%$ and $10\%$ with SegFormer-B2 as the segmentation model
to verify the effectiveness, respectively. We also conduct experiments of OnlySup and FullySup as reference, where OnlySup uses only 1\%, 5\%, and 10\% of labeled data for training, while FullySup employs the entire dataset which represents the 100\% labeled data. The experimental results are presented in Tables \ref{tab:table5} and \ref{tab:table6}, respectively.
\begin{table*}[!t]
\caption{Comparison results with SOTA methods on ISPRS-Potsdam dataset.\ The best results are highlighted in bold. IoU, mIoU, mF1, and Kappa are represented as percentages.\label{tab:table5}}
\centering
\normalsize
\begin{tabular*}{\textwidth}{@{\extracolsep{\fill}} c c c @{\hspace{0.5em}} c @{\hspace{0.75em}} c @{\hspace{1.5em}} c @{\hspace{0.5em}} c @{\hspace{0.5em}} c @{}} 
		\toprule
		\textbf{Ratio} & \textbf{Model} &\multicolumn{5}{c}{\textbf{IoU}}  & \textbf{mIoU}/\textbf{mF1}/\textbf{Kappa} \\
		\cline{3-7}
		~ & ~ & Building & \makecell{Low \\ vegetation}  & Tree & Car & \makecell{Impervious \\ surfaces} &  ~\\
		\midrule
		\multirow{11}{*}{\textbf{1\%}}& OnlySup & 69.80 & 59.39 & 59.50 & 66.50 & 55.85 &  62.21 / 76.58 / 0.6531\\
        ~& Mean teacher\cite{tarvainen2017mean} & 72.53 & 58.98 
         &60.43&67.97&67.39&65.46 / 79.01 / 0.7187\\
        ~& CutMix\cite{yun2019cutmix} & 55.58 & 42.05 &49.72&50.86&39.40&47.52 / 64.21 / 0.5121\\
		~& CCT\cite{ouali2020semi} & 54.48 & 61.28 &48.56&52.95&60.71&55.59 / 71.34 / 0.5761\\
        ~& CPS\cite{chen2021semi} & 59.35 & 69.16 &62.89&59.88&66.33&63.52 / 77.63 / 0.6539\\
		~& LSST\cite{lu2022simple} & 68.74 & 75.24 &54.74&62.09&68.80&65.92 / 79.25 / 0.6847\\
		~& FixMatch\cite{sohn2020fixmatch} & 76.95 & 71.59 &64.71&65.85&72.81&70.38 / 82.53 / 0.7287\\
		~& UniMatch\cite{yang2023revisiting} & 76.52 & 70.99 &\textbf{65.44}&66.62&72.64&70.44 / 82.59 / 0.7291\\
        ~& DWL\cite{huang2024decouple} & 72.34 & \textbf{77.08} &62.74&62.57&72.22&69.39 / 81.79 / 0.7191\\
		~& AllSpark\cite{Wang_2024_CVPR} & 83.70 & 65.92 &59.64&69.77&75.31&70.87 / 82.68 / 0.7827\\
		~& \textbf{Our (MUCA)} & \textbf{84.56} & 66.98 &56.96&\textbf{71.52}&\textbf{76.64}&\textbf{71.33 / 82.92 / 0.7880}\\
		\midrule
		\multirow{11}{*}{\textbf{5\%}}& OnlySup & 84.22 & 64.10&60.81&74.04&76.88&72.01 / 83.10 / 0.7371\\
         ~& Mean teacher\cite{tarvainen2017mean} & 82.15 & 65.92 
         &67.11&72.21&74.60&72.40 / 83.86 / 0.7403\\
		~& CutMix\cite{yun2019cutmix} & 52.94 & 68.86 &41.51&58.33&54.82&55.29 / 70.79 / 0.5783\\
		~& CCT\cite{ouali2020semi} & 72.90 & 80.25 &64.23&58.32&74.42&70.02 / 82.12 / 0.7236\\
        ~& CPS\cite{chen2021semi} & 76.53 & 84.34 &57.98&69.45&75.39&72.74 / 83.78 / 0.7492\\
		~& LSST\cite{lu2022simple} & 69.26 & 84.55 &67.33&67.49&73.86&72.50 / 83.67 / 0.7399\\
        ~& FixMatch\cite{sohn2020fixmatch} & 78.12 & 74.87 &\textbf{68.89}&66.58&75.30&72.75 / 84.15 / 0.7497\\
		~& UniMatch\cite{yang2023revisiting} & 78.24 & 73.59 &67.17&66.64&75.07&72.14 / 83.73 / 0.7432\\
        ~& DWL\cite{huang2024decouple} & 74.81 & \textbf{85.64} &66.38&62.99&75.68&73.10 / 84.22 / 0.7507\\
		~& AllSpark\cite{Wang_2024_CVPR} & 85.57 & 67.62 &60.61&73.48&77.15&72.88 / 84.04 / 0.7989\\
		~& \textbf{Our (MUCA)} & \textbf{88.45} & 69.53 &61.39&\textbf{74.18}&\textbf{79.56}&\textbf{74.62 / 85.15 / 0.8166}\\
		\midrule
		\multirow{11}{*}{\textbf{10\%}}& OnlySup & 83.01 & 69.26 &66.83&75.02&75.47&73.92 / 84.89 / 0.7564\\
          ~& Mean teacher\cite{tarvainen2017mean} & 84.76 & 69.28 
         &68.83&71.66&76.51&74.21 / 85.07 / 0.7578\\
		~& CutMix\cite{yun2019cutmix} & 64.55 & 80.99 &64.79&65.50&68.01&68.77 / 81.34 / 0.7109\\
		~& CCT\cite{ouali2020semi} & 73.09 & 83.94 &61.12&60.45&73.06&70.33 / 82.27 / 0.7265\\
        ~& CPS\cite{chen2021semi} & 77.80 & 87.15 &61.12&68.48&75.89&74.09 / 84.55 / 0.7533\\
		~& LSST\cite{lu2022simple} & 70.92 & 86.06 &68.91&70.22&74.89&74.20 / 84.95 / 0.7549\\
        ~& FixMatch\cite{sohn2020fixmatch} & 77.97 & 76.17 &70.09&70.97&76.14&74.27 / 85.20 / 0.7606\\
		~& UniMatch\cite{yang2023revisiting} & 77.34 & 87.75 &\textbf{70.79}&56.65&76.46&73.80 / 84.52 / 0.7599\\
        ~& DWL\cite{huang2024decouple} & 76.37 & \textbf{88.42} &66.54&64.37&77.14&74.57 / 85.16 / 0.7628\\
		~& AllSpark\cite{Wang_2024_CVPR} & 86.29 & 69.83 &64.17&75.23&78.31&74.76 / 85.35 / 0.8144\\
		~& \textbf{Our (MUCA)} & \textbf{88.02} & 70.58 &64.53&\textbf{75.20}&\textbf{79.92}&\textbf{75.65 / 85.90 / 0.8245}\\
		\midrule
		\textbf{100\%} & FullySup & 90.81 & 72.08 &68.32&76.09&82.56&77.97 / 87.40 / 0.8459\\
		\bottomrule
	\end{tabular*}
     \label{tab_V}
\end{table*}
\begin{table*}[!t]
\caption{Comparison results with other SOTA methods on the LoveDA dataset. The best results are in bold. IoU, mIoU, mF1, and Kappa1 are represented as percentages. \label{tab:table6}}
\centering
\normalsize
\begin{tabular*}{\textwidth}{@{\extracolsep{\fill}} c c c @{\hspace{0.5em}} c @{\hspace{1.5em}} c @{\hspace{2em}} c @{\hspace{1.5em}} c @{\hspace{1.5em}} c @{\hspace{0.5em}} c @{\hspace{0.5em}} c @{}} 
		\toprule
		\textbf{Ratio} & \textbf{Model} &\multicolumn{7}{c}{\textbf{IoU}}  & \textbf{mIoU}/\textbf{mF1}/\textbf{Kappa} \\
		\cline{3-9}
		~ & ~ & Background &  Building &Road & Water & Barren & Forest & Agriculture &  ~\\
		\midrule
	\multirow{11}{*}{\textbf{1\%}}& OnlySup & 41.96 & 37.86 &36.74&57.03&9.02&31.70&37.73&36.01 / 51.41 / 0.4141\\
        ~& Mean teacher\cite{tarvainen2017mean} & 44.73 & 42.53 &40.34&50.92&11.37&26.88&54.40&38.73 / 52.81 / 0.4251\\
		~& CutMix\cite{yun2019cutmix} & 36.04 & 24.69 &10.03&24.60&3.43&6.67&10.19&16.52 / 26.85 / 0.1362\\
		~& CCT\cite{ouali2020semi} & 37.16 & 22.41 &27.86&43.98&14.51&25.38&36.67&29.71 / 44.99 / 0.3517\\
        ~& CPS\cite{chen2021semi} & 46.52 & 20.87 &27.85&50.55&0.01&33.16&34.60&30.51 / 44.28 / 0.3802\\
		~& LSST\cite{lu2022simple} & 44.73 & 41.90 &39.90&62.65&29.27&31.26&48.29&42.57 / 59.00 / 0.4817\\
        ~& FixMatch\cite{sohn2020fixmatch} & 46.78 & 51.20 &50.21&67.27&11.53&\textbf{36.79}&50.26&44.86 / 60.02 / 0.5175\\
		~& UniMatch\cite{yang2023revisiting} & 46.53 & 51.38 &49.36&\textbf{67.74}&10.86&33.40&52.28&44.51 / 59.51 / 0.5175\\
		~& DWL\cite{huang2024decouple} & 48.74 & 56.79 &\textbf{51.59}&63.42&22.56&35.20&55.38&47.67 / 63.40 / 0.5534\\
        ~& AllSpark\cite{Wang_2024_CVPR} & 63.87 & 47.70 &46.05&61.52&35.31&30.94&55.64&48.72 / 63.29 / 0.5502\\
		~& \textbf{Our (MUCA)} & \textbf{64.89} & \textbf{56.03} &47.14&63.86&\textbf{35.81}&22.57&\textbf{58.18}&\textbf{49.78 / 63.48 / 0.5753}\\
		\midrule
	\multirow{11}{*}{\textbf{5\%}}& OnlySup & 48.00 & 43.34 &50.56&61.42&23.14&\textbf{37.63}&45.00&44.20 / 60.46 / 0.5061\\
      ~& Mean teacher\cite{tarvainen2017mean} & 49.73 & 46.22 &42.34&60.93&31.51&35.79&44.22&44.39 / 61.81 / 0.5151\\
		~& CutMix\cite{yun2019cutmix} & 41.48 & 41.62 &38.77&47.44&14.69&28.09&31.05&34.73 / 50.65 / 0.3692\\
		~& CCT\cite{ouali2020semi} & 46.80 & 44.62 &46.80&60.95&24.83&29.03&44.30&42.48 / 58.74 / 0.4850\\
        ~& CPS\cite{chen2021semi} & 48.90 & 49.64 &47.97&60.27&4.67&36.09&47.32&42.12 / 56.90 / 0.4976\\
		~& LSST\cite{lu2022simple} & 51.48 & 45.66 &52.66&67.63&33.52&35.80&48.60&47.91 / 64.10 / 0.5434\\
        ~& FixMatch\cite{sohn2020fixmatch} & 45.40 & 53.05 &\textbf{51.22}&66.73&28.53&27.25&54.30&44.64 / 62.45 / 0.5378\\
		~& UniMatch\cite{yang2023revisiting} & 50.20 & 54.49 &50.46&67.18&26.79&30.06&54.86&47.72 / 63.46 / 0.5543\\
        ~& DWL\cite{huang2024decouple} & 48.75 & 55.00 &51.53&\textbf{69.49}&29.46&36.59&52.11&48.99 / 64.88 / 0.5597\\
		~& AllSpark\cite{Wang_2024_CVPR} & 65.09 & 55.06 &47.59&67.10&34.67&26.86&51.87&49.75 / 64.91 / 0.5682\\
		~& \textbf{Our (MUCA)} & \textbf{67.29} & \textbf{56.04} &48.37&61.02&\textbf{36.21}&30.76&\textbf{57.09}&\textbf{50.97 / 64.92 / 0.5856}\\
		\midrule
		\multirow{11}{*}{\textbf{10\%}}& OnlySup & 47.02 & 51.04 & 50.73& 59.42& 31.66&38.25&51.26& 47.05 / 63.52 / 0.5254\\
         ~& Mean teacher\cite{tarvainen2017mean} & 50.45 & 55.75 &43.56&66.15&35.24&36.96&45.64&47.68 / 64.18 / 0.5387\\
		~& CutMix\cite{yun2019cutmix} & 46.73 & 49.60 &47.36&59.99&29.06&37.77&40.60&44.44 / 60.99 / 0.4837\\
		~& CCT\cite{ouali2020semi} & 44.07 & 45.22 &47.65&57.12&24.41&32.50&45.07&42.29 / 58.73 / 0.4762\\
        ~& CPS\cite{chen2021semi} & 51.30 & 54.93 &52.57&53.37&18.39&37.59&53.24&45.91 / 61.78 / 0.5479\\
		~& LSST\cite{lu2022simple} & 50.69 & 49.50 &52.63&69.85&27.25&36.24&52.06&48.32 / 64.17 / 0.5565\\
        ~& FixMatch\cite{sohn2020fixmatch} & 52.02 &  55.59 &53.20& 57.91& 25.86&40.83& 57.50& 48.99 / 64.97 / 0.5676\\
		~& UniMatch\cite{yang2023revisiting} & 51.80 & 53.95 & 51.17&58.15&25.60&38.72&54.86&47.75 / 63.86 / 0.5639\\
        ~& DWL\cite{huang2024decouple} & 49.94 &  56.66 & \textbf{53.89}&\textbf{70.35}& 30.62&\textbf{41.49}&53.13&50.87 / 66.64 / 0.5753\\
		~& AllSpark\cite{Wang_2024_CVPR} & 67.13 & 56.16 &40.67&63.58&32.54&32.03&56.91&49.86 / 63.97 / 0.5751\\
		~& \textbf{Our (MUCA)} & \textbf{68.69} & \textbf{58.20} &41.82&65.62&\textbf{37.09}&35.01&\textbf{57.38}&\textbf{51.97 / 66.72 / 0.5901}\\
		\midrule
		\textbf{100\%} & FullySup & 68.84 & 58.57 &48.02&70.39&43.28&38.59&62.30&55.71 / 69.10 / 0.6291\\
		\bottomrule
	\end{tabular*}
    \label{tab_VI}
\end{table*}
From the viewpoint of methods, our observations reveal the following trends. Methods such as CutMix and CCT exhibit a performance decline when incorporating unlabeled data into the training process, compared to the basic OnlySup method. In contrast, methods like FixMatch, UniMatch, DWL, and the proposed MUCA demonstrate a beneficial impact on performance. MUCA achieves the best results. There are many possible reasons. For example, methods like CutMix and CCT use all unlabeled data for consistency regularization during training without selective filtering. This may inadvertently result in memorizing a large number of incorrect pseudo-labels. On the other hand, FixMatch, UniMatch, DWL, and MUCA employ strategies such as careful filtering or weighting of pseudo-labels in the unlabeled data, which reduces the harmful impact of low-confidence samples and prevents the dominance of most classes. 

Additionally, an interesting phenomenon is that sometimes the performance of UniMatch is lower than that of FixMatch. UniMatch extends FixMatch by introducing an additional perturbation branch and a feature re-perturbation branch to enrich the representation space. Both branches are supervised by a weakly augmented branch. However, RS datasets often contain too many hard-to-distinguish classes and noise. They hinder the training process on the unlabeled data, which could be exacerbated by the inclusion of additional pseudo-supervision branches. In contrast, our method achieves better results by using MSUC to exclude pseudo-labels with high uncertainty and utilizing CTSA to enhance the correlation WA and SA.

Furthermore, it is worth noting that our model outperforms all other SOTA methods on ISPRS-Potsdam and LoveDA datasets when the labeled data ratios are 1\%, 5\%, and 10\%, respectively. This is a good demonstration of the fact that our semi-supervised model achieves better applicability on RS datasets with a very small labeled data ratio.

For specific classes, our model performs well from expansive classes to small classes, such as large-scale Building and small-scale Car‌‌. Our model achieves SOTA performance on Building which is a class with extreme scale variations. Both datasets include categories that are difficult to distinguish. For example, the ISPRS-Potsdam dataset includes Building, Low vegetation, Tree, and Impervious surfaces, while the LoveDA dataset includes categories such as Building, Barren, Forest, and Agriculture. More specific, classes including Building‌, Impervious surfaces‌,‌ Barren‌, Road‌, and Background‌ predominantly exhibit‌ grayish tones‌ and may share  similar rough or smooth textures‌. Meanwhile, ‌Low vegetation‌, Tree‌, Forest‌, and Agriculture‌, which predominantly appear in green hues‌, also pose segmentation challenges for models‌. Our method, MUCA, achieves an improvement on IoU for almost all classes compared to OnlySup when the labeled data ratios are 1\%, 5\%, and 10\%. Additionally, when the labeled data ratios are 1\%, 5\%, and 10\%, MUCA outperforms other SOTA models in $IoU$ performance for the Building, Car, and Impervious surfaces classes in the ISPRS-Potsdam dataset, and for the Background, Building, Barren, and Agriculture classes in the LoveDA dataset. Although MUCA does not achieve the best $IoU$ for all classes, this phenomenon is understandable and explainable. This is because different models have different focuses, which make one algorithm perform exceptionally well on specific classes while underperforming for others. However, focusing solely on improving a single class is not the research focus of this paper. For example, while the DWL model achieves significant improvement in the Low vegetation class, its performance on the Building, Car, and Impervious surfaces classes is unsatisfactory.

\subsection{Model generalizability experiments}
The proposed model is non-intrusive and can be easily integrated into existing semantic segmentation networks without changing the network structure itself. Therefore, we conduct experiments by integrating our model into several popular networks including U-Net, PSPNet, DeepLabv3+, and SegFormer-B2 on RS datasets. These experiments are designed to evaluate the general applicability of the MUCA on segmentation models. The first three of these models are classical CNN semantic segmentation models, and the last one is based on the Transformer architecture. We comparative analyze the performance improvement of MUCA with Onlysup and NoUC on these models. NoUC, as clearly defined in the Ablation Study section, refers to a simplified method that does not perform multi-scale uncertainty estimation and not incorporate the CTSA module. Instead, it relies solely on standard consistency regularization to calculate the loss of feature maps at each stage. The model general applicability experiments were performed on 5\% labeled training data and the results are shown in Table \ref{tab:table7}.

\begin{table}[!t]
\caption{Model generalizability experiments. \label{tab:table7}}
\centering
\normalsize
\begin{tabular}{ c |c |c |c}
		\toprule
		\textbf{Dataset} & \textbf{Network}  & \textbf{Model}&\textbf{mIoU($5\%$)} \\
		\midrule
		 \multirow{12}{*}{ISPRS-Potsdam} & \multirow{3}{*}{U-Net} & Onlysup & 57.52  \\ 
          ~ &~ & NoUC & 61.59  \\
            ~ &~ & MUCA & 64.21  \\
        \cmidrule{2-4}
		 ~ &\multirow{3}{*}{PSPNet} & Onlysup & 63.46  \\ 
        ~ & ~ & NoUC & 66.16  \\
          ~ & ~ & MUCA & 68.87  \\
		\cmidrule{2-4}
		 ~ &\multirow{3}{*}{DeepLabv3+} & Onlysup & 66.83  \\ 
         ~ &~ & NoUC & 68.01  \\
         ~ &  ~ & MUCA & 70.49  \\
		\cmidrule{2-4}
		~ & \multirow{3}{*}{SegFormer-B2} & Onlysup &72.01  \\ 
         ~ &  ~ & NoUC & 73.14  \\
         ~ &  ~ & MUCA & 74.62  \\
         \midrule
          \multirow{12}{*}{LoveDA} & \multirow{3}{*}{U-Net} & Onlysup & 38.12  \\ 
          ~ &~ & NoUC & 40.63  \\
            ~ &~ & MUCA & 42.32  \\
        \cmidrule{2-4}
		 ~ &\multirow{3}{*}{PSPNet} & Onlysup & 40.16  \\ 
        ~ & ~ & NoUC & 42.98  \\
          ~ & ~ & MUCA & 44.56  \\
		\cmidrule{2-4}
		 ~ &\multirow{3}{*}{DeepLabv3+} & Onlysup & 41.45  \\ 
         ~ &~ & NoUC & 43.87  \\
         ~ &  ~ & MUCA & 45.98  \\
		\cmidrule{2-4}
		~ & \multirow{3}{*}{SegFormer-B2} & Onlysup & 44.20  \\ 
           ~ &  ~ & NoUC & 48.21  \\
         ~ &  ~ & MUCA & 50.97  \\
		\bottomrule
	\end{tabular}
\end{table}

It can be seen that semi-supervised model MUCA achieves significant improvements with 5\% labeled training data. It enhances the performance of popular semantic segmentation models including U-Net, PSPNet, DeepLabv3+, and SegFormer-B2. This shows that our MUCA has the ability to improve the performance of classical CNN-baed semantic segmentation models and novel Transformer architecture models.

\subsection{Visualization Comparison}
\begin{figure*}[!t]
\centering
\includegraphics[width=7in]{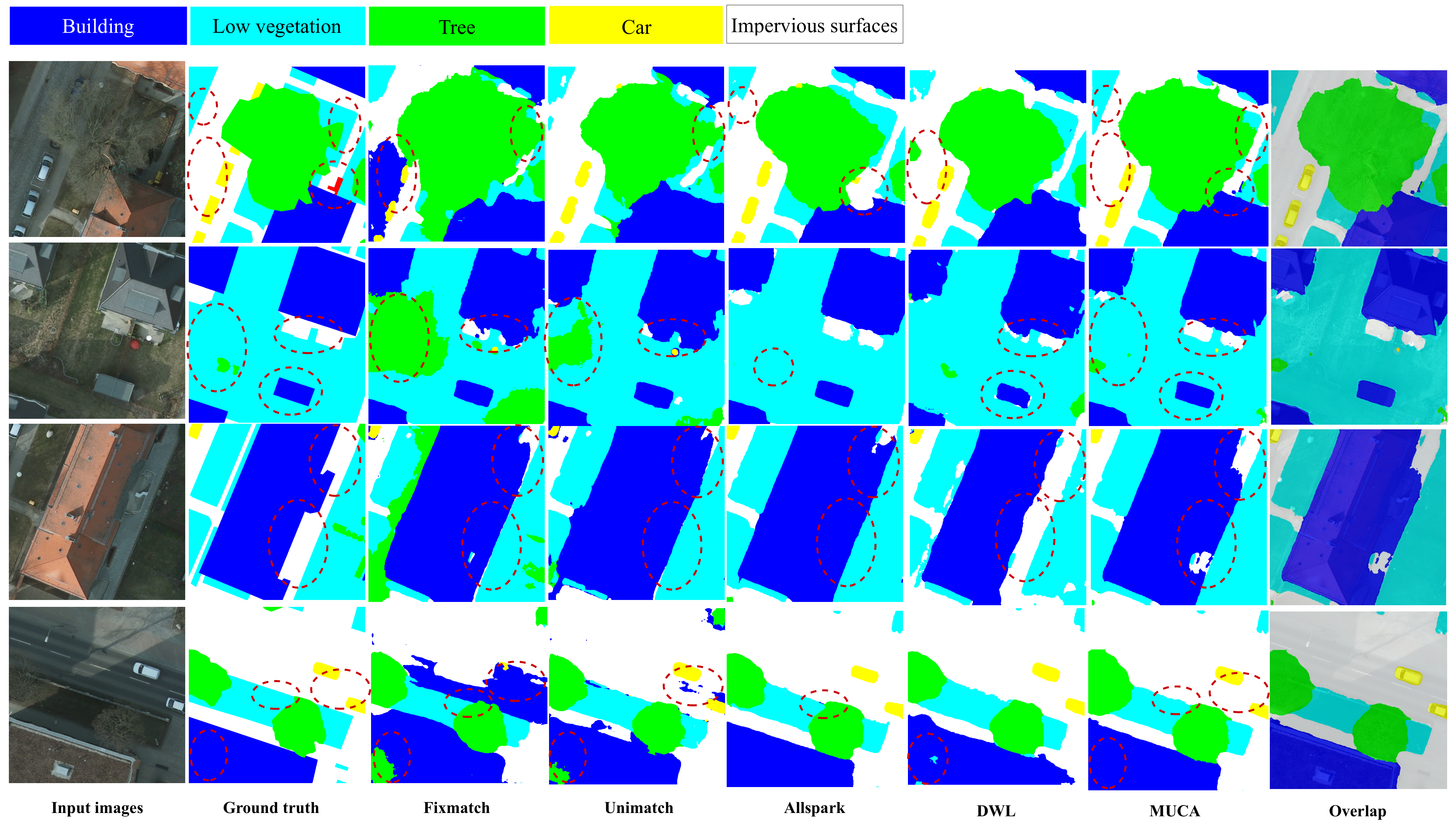}\vspace{-0.2cm}
\caption{Visual comparison of semantic segmentation results with different semisupervised methods on the ISPRS-Potsdam dataset.
\label{Potsdam}\vspace{-0.5cm}}
\end{figure*}

\begin{figure*}[!t]
\centering
\includegraphics[width=7in]{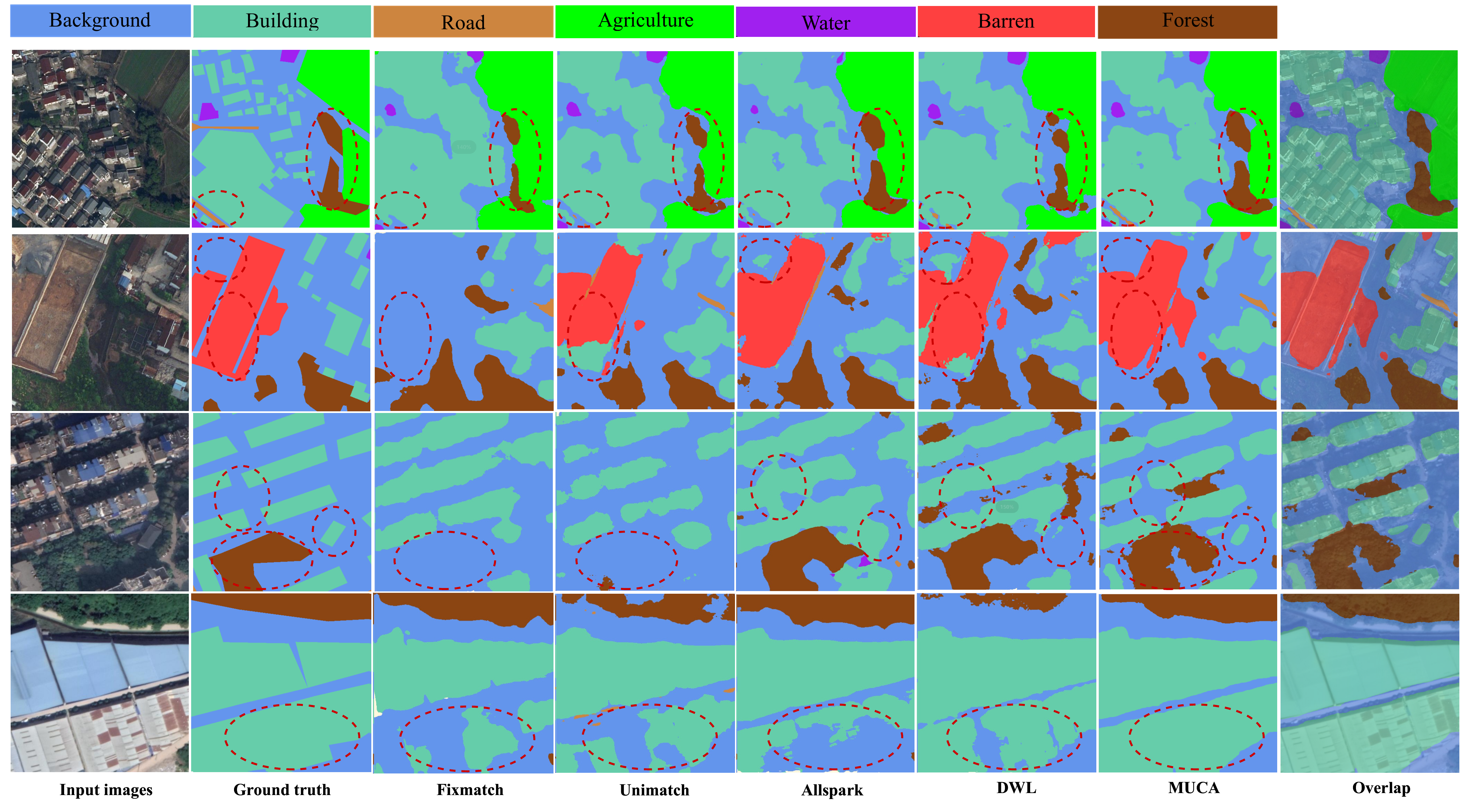}\vspace{-0.2cm}
\caption{Visual comparison of semantic segmentation results with different semisupervised methods on the LoveDA dataset.
\label{LoveDA}
}\vspace{-0.5cm}
\end{figure*}

We conducted visual comparison experiments to see the advantages of our approach more intuitively and clearly. Fig. \ref{Potsdam} and Fig. \ref{LoveDA} show the visual comparison of several semi-supervised semantic segmentation methods on the ISPRS-Potsdam dataset and LoveDA dataset. Additionally, we have highlighted the mis-segmented areas of other SOTA methods with dashed ellipses, while emphasizing the comparatively accurate segmentation results of our approach in the same regions for contrast.

For ISPRS-Potsdam dataset, the Fixmatch and Unimatch make errors in segmenting regions for the Low vegetation, Tree, Impervious surfaces and Building classes. Allspark and DWL successfully recognize part of the region where Low vegetation classes are mixed with Tree classes, but the segmentation results for Building class exhibit inaccuracies. In contrast, our method shows significant advantages in accurately recognizing and segmenting Building (blue), Car (yellow), and Tree (green). Especially, we can see that methods including Fixmatch and Unimatch show large-scale segmentation errors when Car, Low vegetation and Tree are mixed. In addition, we overlap our segmentation results with the original image in the last column to clearly show our advantage. This overlapped image demonstrates that our model achieves excellent visual results, particularly in the Tree and Car categories.

The objective metrics of all models on the LoveDA dataset are lower compared to those on the ISPRS-Potsdam dataset. Consequently, the visualization effects of all models become less effective compared to their performance on the ISPRS-Potsdam dataset. Specifically, FixMatch and UniMatch exhibit noticeable errors in segmentation areas of Building, Barren, and Background classes. While Allspark and DWL models achieve relatively precise segmentation boundaries, they still demonstrate poor classification performance in mixed zones containing both Building and Barren classes‌. Furthermore, the segmentation accuracy for individual structures within the Building class remains suboptimal, failing to delineate architectural details effectively. In contrast, our proposed MUCA shows significant advantages in accurately identifying Building, Agriculture, Barren and Background classes. Finally, we overlay the segmentation results with the original images in the last column to clearly demonstrate our advantages. This overlayed visualization indicates that our model achieves superior visual results on the LoveDA dataset compared to other models.

\section{Discussion and Conclusion}
\label{Conclusion}
Our study addresses the challenges associated with semi-supervised RS image semantic segmentation by proposing the Multi-Scale Uncertainty and Cross-Teacher-Student Attention (MUCA) model. The MUCA model includes two special modules, i.e., Multi-scale Uncertainty Consistency (MSUC) and Cross-Teacher-Student Attention (CTSA). The goal of MSUC is to learn rich multi-scale information, meanwhile, CTSA distinguishes the high inter-class similarities through the cross-network attention mechanism. The new features constructed by CTSA enable the student network decoder to benefit from the dual enhancement of both WA and SA, achieving better optimization results and demonstrating greater stability during the training phase.

Compared to SOTA algorithms, our method achieved the best results for the metrics $mIoU$, $mF1$, and $Kappa$, however, did not achieve the best IoU for all the categories. This phenomenon is understandable and explainable. On the one hand, this paper does not address other problems in the semi-supervised domain such as long-tailed distribution among classes. On the other hand, different models have different focuses which may lead to a particularly good result for one class and a poor result for others. This study highlights the more realistic scenario where MUCA demonstrates the best overall performance across all classes. However, current semi-supervised models for RS image analysis still face several bottlenecks. For instance, segmentation performance remains suboptimal in areas with mixed land cover, and visualized results on the LoveDA dataset exhibit flawed boundaries for classes such as ‌Forest and ‌Road‌. Notably, neither our proposed ‌MUCA‌ nor existing semi-supervised models achieve an mIoU exceeding 52\% on LoveDA, indicating substantial room for improvement. Furthermore, research on semi-supervised models for complex spectral imaging (e.g., multispectral) and multimodal RS data fusion remains limited. Our future work will focus on ‌cross-modal alignment and developing semi-supervised models that integrate multi-source data, including multispectral imagery, SAR, and LiDAR.

Finally, we hope that the model proposed in this paper can serve as a simple and powerful baseline in the field of semi-supervised semantic segmentation of RS images. We also aim to inspire more valuable research for future work.
\section{Acknowledgment}
We would like to express our sincere appreciation to the anonymous reviewers for their insightful comments, which have greatly aided us in improving the quality of the paper.
\bibliographystyle{IEEEtran}
\bibliography{muca}

\begin{IEEEbiography}[{\includegraphics[width=0.9in,height=1.25in,clip,keepaspectratio]{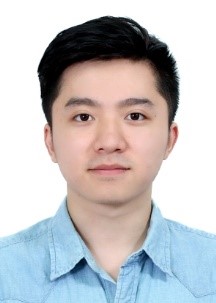}}]{Shanwen Wang} received the M.Sc. degree from Chengdu University of Technology, Chengdu, China, in 2023. He is currently pursuing the Ph.D. degree with the City University of Macau, Macau, China. His research interests include remote sensing image processing and computer vision.
\end{IEEEbiography}
\vspace{-15 mm}

\begin{IEEEbiography}[{\includegraphics[width=1in,height=1.25in,clip,keepaspectratio]{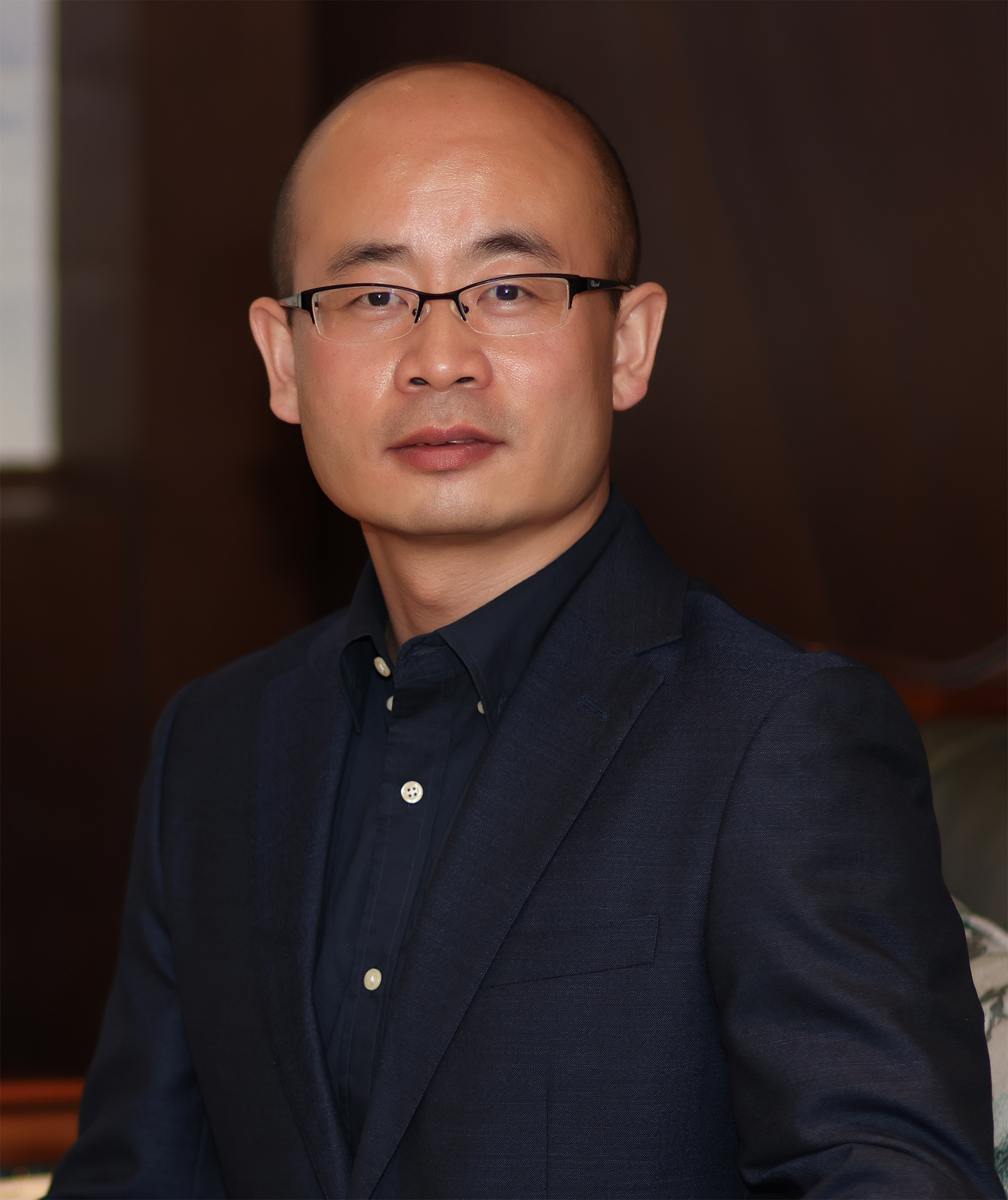}}]{Xin Sun} (Senior Member, IEEE) is a full Professor with Faculty of Data Science at City University of Macau. He was an experienced Humboldt Researcher (2022-2023) at Technical University of Munich (TUM), Munich, Germany. He received his Bachelor's degree, Master (M.Sc.) degree, Ph.D degree from the College of Computer Science and Technology at Jilin University in 2007, 2010 and 2013, respectively. He did the Post-Doc research (2016-2017) in the department of computer science at the Ludwig-Maximilians-Universit\"{a}t M\"{u}nchen, Germany. His current research interests include machine learning, remote sensing and computer vision.
\end{IEEEbiography}
\vspace{-15 mm}
\begin{IEEEbiography}[{\includegraphics[width=0.9in,height=1.25in,clip,keepaspectratio]{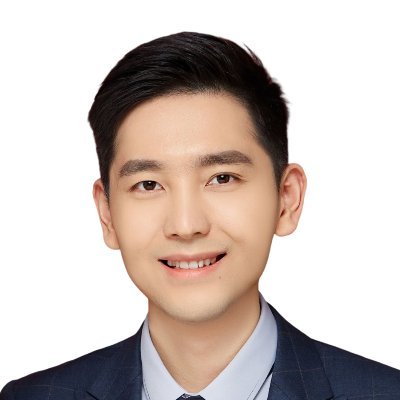}}]{Changrui Chen} received his PhD degree from the WMG, University of Warwick, U.K, and the bachelor's and master's degrees in computer science and technology from the Ocean University of China, in 2017 and 2020, respectively.  His research interests include computer vision and semi-supervised learning.
\end{IEEEbiography}
\vspace{-15 mm}
\begin{IEEEbiography}[{\includegraphics[width=1in,height=1.25in,clip,keepaspectratio]{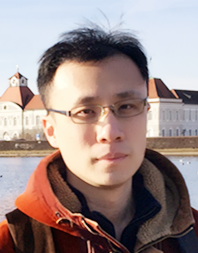}}]{Danfeng Hong} (Senior Member, IEEE) received the Dr.-Ing. degree (summa cum laude) from the Signal Processing in Earth Observation (SiPEO), Technical University of Munich (TUM), Munich, Germany, in 2019. Since 2022, he has been a Full Professor with the Aerospace Information Research Institute, Chinese Academy of Sciences, Beijing, China. His research interests include artificial intelligence, multimodal big data, foundation models, and Earth observation.
Dr. Hong has received several prestigious awards, including the Jose Bioucas Dias Award in 2021 and the Paul Gader Award in 2024 at WHISPERS for outstanding papers, the Remote Sensing Young Investigator Award in 2022, the IEEE GRSS Early Career Award in 2022, and the “2023 China’s Intelligent Computing Innovators” Award (the only recipient in AI for Earth Science) by MIT Technology Review in 2024. He has been recognized as a Highly Cited Researcher by Clarivate Analytics in 2022, 2023, and 2024. He serves as an Associate Editor for IEEE Transactions on Image Processing (TIP) and IEEE Transactions on Geoscience and Remote Sensing (TGRS). He is also an Editorial Board Member of Information Fusion and ISPRS Journal of Photogrammetry and Remote Sensing.
\end{IEEEbiography}
\vspace{-15 mm}
\begin{IEEEbiography}[{\includegraphics[width=1in,height=1.25in,clip,keepaspectratio]{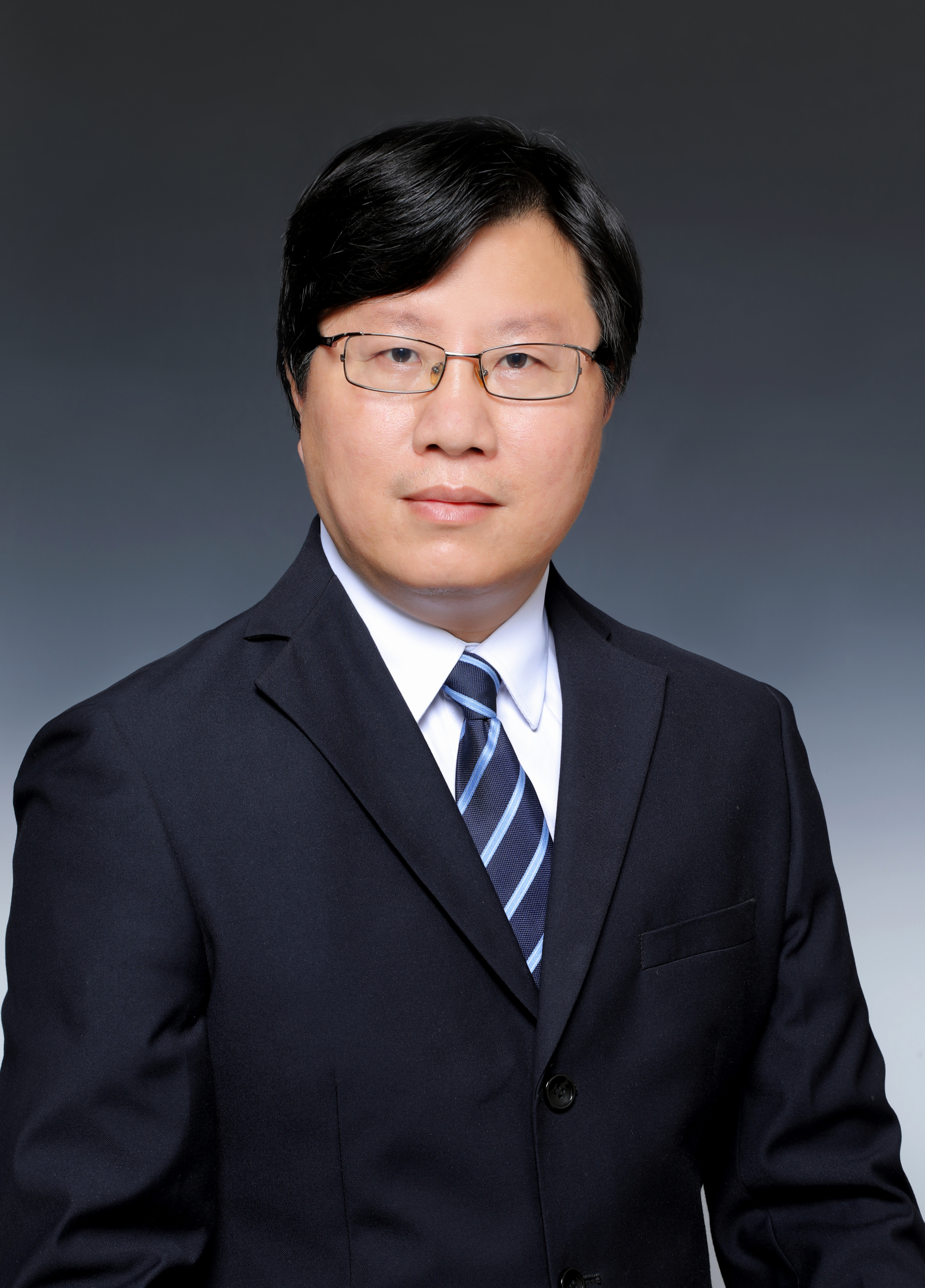}}]{Jungong Han} (Senior Member, IEEE) is with the Beijing National Research Center for Information Science and Technology (BNRist) and the Department of Automation, Tsinghua University, Beijing, China. His research interests include computer vision, artificial intelligence, and machine learning. He is a fellow of the International Association of Pattern Recognition. He serves as an Associate Editor for many prestigious journals, such as IEEE TRANSACTIONS ON NEURAL NETWORKS AND LEARNING SYSTEMS, IEEE TRANSACTIONS ON CIRCUITS AND SYSTEMS FOR VIDEO TECHNOLOGY, IEEE TRANSACTIONS ON MULTIMEDIA, and Pattern Recognition.
\end{IEEEbiography}
\vspace{-15 mm}
\newpage
\vfill

\end{document}